\documentclass[10pt,journal,compsoc]{IEEEtran}

\usepackage{booktabs} 
\usepackage{mathptmx}
\usepackage{graphicx}
\usepackage{amsmath}
\usepackage{caption}
\usepackage{times}
\usepackage{epsfig}
\usepackage{comment}
\usepackage{xspace}
\usepackage{subfig}
\usepackage{url}
\usepackage{makecell}
\usepackage{xcolor}
\usepackage[backend=biber,style=numeric,sorting=ynt]{biblatex}
\usepackage{multirow}
\usepackage{multicol}
\robustify{\subref}
\newcommand{\IGNORE}[1]{}
\usepackage{array}

\newcommand{\observations}[1]{%
\vspace{0.15cm plus 0cm minus 0.1cm}
\hspace{-0.45cm}%
\fbox{%
\begin{minipage}{0.96\linewidth}
    #1
\end{minipage}%
}%
}

\begin{document}

\title{
The Life and Death of SSDs and HDDs: Similarities, Differences, and Prediction Models
}

\author{
        Riccardo~Pinciroli,
        Lishan~Yang,
        Jacob~Alter,
        and~Evgenia~Smirni
        \IEEEcompsocitemizethanks{
            \IEEEcompsocthanksitem R. Pinciroli, L. Yang,  E. Smirni are with the Computer Science Department, William and Mary, Williamsburg, VA 23185.\newline E-mail: \{rpinciro,lyang11,esmirni\}@cs.wm.edu \newline
            J. Alter is with Ipsos North America. E-mail: 
            jralter@email.wm.edu
        }
}

\IEEEtitleabstractindextext{
    \begin{abstract}
    Data center downtime typically centers around IT equipment failure. Storage devices are the most frequently failing components in data centers.
    We present a comparative study of hard disk drives (HDDs) and solid state drives (SSDs) that constitute the typical storage in data centers.
    Using a six-year field data of 100,000 HDDs of different models from the same manufacturer from the BackBlaze dataset and a six-year field data of 30,000 SSDs of three models from a Google data center, we characterize the workload conditions that lead to failures and illustrate that their root causes differ from common expectation but remain difficult to discern.
    For the case of HDDs we observe that young and old drives do not present many differences in their failures. Instead, failures may be distinguished by discriminating drives based on the time spent for head positioning.
    For SSDs, we observe high levels of infant mortality and characterize the differences between infant and non-infant failures. We develop several machine learning failure prediction models that are shown to be surprisingly accurate, achieving high recall and low false positive rates. These models are used beyond simple prediction as they aid us to untangle the complex interaction of workload characteristics that lead to failures and identify failure root causes from monitored symptoms.
    \end{abstract}
    
    \begin{IEEEkeywords}
    \textbf{Supervised learning, Classification, Data centers, Storage devices, SSD, HDD}
    \end{IEEEkeywords}
}

\maketitle

\IEEEdisplaynontitleabstractindextext

\maketitle

\section{Introduction}
\label{sec:intro}

Storage devices, such as hard disk drives (HDDs) and solid state drives (SSDs), are among the components that affect the data center dependability the most \cite{schroeder2007disk,schroeder2009dram,hwang2012cosmic,DBLP:conf/fast/BirkeBCSE14}, contributing to the 28\% of data center failure events \cite{xu2019lessons}.
Accurate prediction of storage component failures enables on-time mitigation actions that avoid data loss and increases data center dependability.
Failure events observed in HDDs and SSDs are different due to their distinct physical mechanics, 
it follows that observations from HDD analysis cannot be generalized to SSDs and vice-versa.
Existing studies of HDD dependability in the literature generally focus on a few disk models that are analyzed using disk traces of collected disk data during a period of up to two years  \cite{botezatu2016predicting,wang2019attention}.
Most of the SSD studies in the literature are done on a simulated or controlled environment \cite{DBLP:conf/micro/GruppCCSYSW09,DBLP:conf/fast/ZhengTQL13,DBLP:conf/hpca/CaiLHMM15,DBLP:conf/dsn/QureshiKKNM15,DBLP:conf/sigmetrics/LuoGCHM18} or focus on specific error types \cite{MezaWKM15,mahdisoltani2017proactive,schroeder2017reliability}.

In this paper, we focus on failures of HDDs and SSDs by analyzing disk logs collected from real-world data centers over a period of six years.
We analyze \textit{Self-Monitoring, Analysis and Reporting Technology} (SMART) traces from five different HDD models from the Backblaze data center \cite{backblaze} and the logs of three multi-level cell (MLC) models of SSDs collected at a Google data center.
Though we are unaware of the data centers' exact workflows for drive repairs, replacements, and retirements (e.g., whether they are done manually or automatically, or the replacement policies in place), we are able to discover key correlations and patterns of failure, as well as generate useful forecasts of future failures.
Being able to predict an upcoming drive retirement could allow early action: for example,
early replacement before failure happens, migration of data and VMs to other resources, or even allocation of VMs to disks that are not prone to failure~\cite{XuATC18}.

In this paper, we study the various error types accounted by the logs to determine their roles in triggering, or otherwise portending, future drive failures.
It is interesting to note that although both the Backblaze and the Google logs have ample data, simple statistical methods are not able to achieve highly accurate predictions: we find no evidence that the failure/repair process is triggered by any deterministic decision rule. 
Since the complexity of the data does not allow for a typical treatment of prediction based on straightforward statistical analysis, we resort to machine learning predictors to help us detect which quantitative measures provide strong indications of upcoming failures.
We show that machine learning models that are trained from monitoring logs achieve failure prediction that is both
remarkably accurate and timely, both in the HDD and SSD domains.
The models confirm that drive replacements are not triggered by naive thresholding of predictors. 
Beyond prediction, the models are interpreted to provide valuable insights on which errors and workload characteristics are most indicative of future catastrophic failures.

The models are able to anticipate failure events with reasonable accuracy up to several days in advance, despite some inherent methodological challenges. Although the frequency of failures is significant, the datasets in hand are highly imbalanced (the healthy-faulty ratio is 10,000:1).
This is a common problem in classification, and makes achieving simultaneously high true positive rates and low false positive ones very difficult. 
Here, we illustrate that there exist different ways to partition the HDD and SSD datasets to increase model accuracy.
This partitioning is based on  
 workload analysis that was first developed in  \cite{alter2019ssd} for SSDs and focuses on the discovery of certain drive attributes. We saw that similar partitioning can be also successfully applied for the case of HDDs.

We focus on the interpretability of the machine learning models and derive insights 
that can be used to drive proactive disk management policies.
Our findings are summarized as follows:
\begin{itemize}
    \item Although a consistent number of drives fail during the observed six years (i.e., 14\% of SSDs and 7\% of HDDs), only a few of them are repaired and put back into production (i.e., 50\% of failed SSDs and 7\% of failed HDDs).
    The percentage of failed SSDs that are put back into production within a month after the failure event is 8\%, while almost all the repaired HDDs are back in the data center within a day from the failure.
    \item Drive failures are triggered by a set of attributes and different drive features must be monitored to accurately predict a failure event. There is no single metric that triggers a drive failure after it reaches a certain threshold.
    \item Several machine learning predictors are  quite successful for failure prediction. Random forests are found to be the most successful for SSDs and HDDs. 
    \item Datasets may be partitioned to improve the performance of the classifier.
    Partitioning SSDs on the \textit{Drive age} attribute and HDDs on \textit{head flying hours} (i.e., SMART 240) increases model accuracy.
    This is enabled by training a distinct classifier for each partition of the dataset.
    \item The attributes that are most useful for failure prediction differ depending on the dataset split (i.e., split based on age for SSDs and on head flying hours for HDDs).
    \item No relationship between SSD program-erase (P/E) cycles and failures are observed, suggesting that that write operations do not affect the state of SSDs as much as previously thought.
\end{itemize}

The remainder of the paper is organized as follows.
Sections \ref{sec:dataset} describes the datasets analyzed in this paper and Section \ref{sec:characterization} characterizes the data and summarizes observations.
Section \ref{sec:causes} investigates main reasons causing SSD and HDD failures.
Section \ref{sec:prediction} studies different classification algorithm to predict failure events and presents a partition strategy to increase the performance of models.
Section \ref{sec:related} presents related work and Section \ref{sec:conclusion} concludes the paper.
\section{SSD and HDD Dataset}
\label{sec:dataset}

\begin{figure}
\centering
    \begin{minipage}{0.49\columnwidth}
 		\centering
  		\includegraphics[scale=0.56]{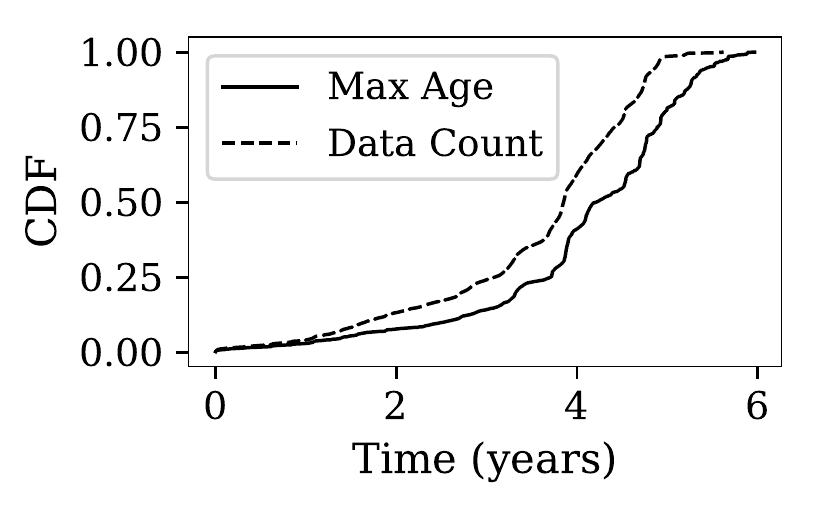}
  		
  		(a) SSD
 	\end{minipage}
 	 	\begin{minipage}{0.49\columnwidth}
 		\centering
  		\includegraphics[scale=0.56]{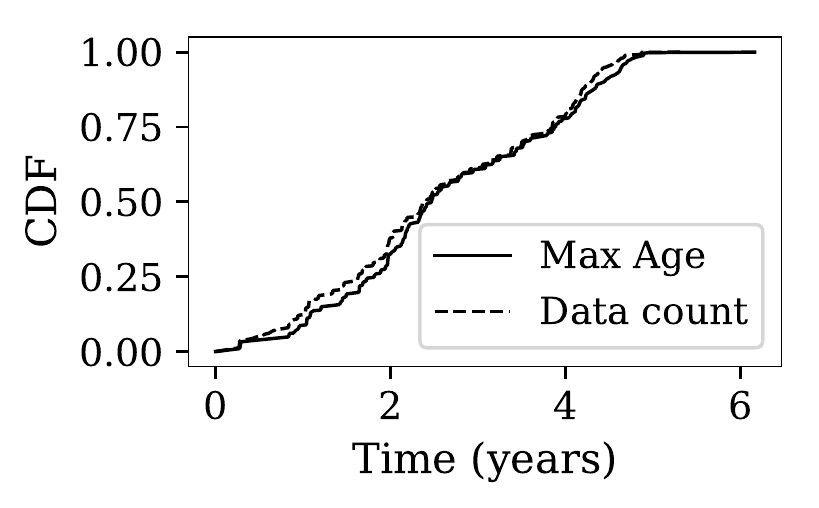}
  		
  		(b) HDD
 	\end{minipage}

\caption{
CDFs of maximum observed drive age (solid) and number of observed drive days within the error logs (dashed).
The HDD maximum age is shorter than the SSD one.
}
\label{fig:agecdf}
\end{figure}

\begin{table*}
	\centering
    \caption{Proportion of drive days that exhibit each error type.}
    \label{table:different-error}
	\centering
    \begin{tabular}{c|l|lllllllll} \toprule
\multirow{5}{*}{SSD} & \multirow{2}{*}{Error Type} & {correctable }   & {final read } & {final write } & {meta } & {read } & {response }  & {timeout } & {uncorrectable } & {write } \\ 
     & & {error} & {error} & {error} & {error} & {error} & {error} & {error} & {error} & {error}  \\ \cline{2-11}
     & MLC-A & 0.828895 & 0.001077 & 0.000026 & 0.000014 & 0.000090 & 0.000001 & 0.000009 & 0.002176 & 0.000117  \\ 
     & MLC-B & 0.776308 & 0.001805 &  0.000027 & 0.000016 &  0.000103 & 0.000004 & 0.000010 &  0.002349 & 0.001309   \\ 
     & MLC-D & 0.767593 & 0.001552  & 0.000034  & 0.000028 &  0.000133  & 0.000002  & 0.000014 & 0.002583 &  0.000162  \\ \midrule
     	\end{tabular}
	\begin{tabular}{c|l|lllll} 
\multirow{7}{*}{HDD} &    \multirow{2}{*}{Error Type~~~~~~~~~~~~~~~~~~~~}          & uncorrectable~~~~ & ultraDMC CRC~~~~  & reallocated sector~~~~ & power-off retract~~~~ & uncorrectable sector  \\ 
 &   & error &  error & count & count & count\\ 
\cline{2-7}
                     & ST12000NM0007 & 0.004301            & 0.007232           & 0.014590           & 0.330956          & 0.003755             \\
                     & ST3000DM001   & 0.216522            & 0.303442           & 0.092016           & 0.001818          & 0.074695             \\
                     & ST4000DM000   & 0.014837            & 0.022238           & 0.004435           & 0.000549          & 0.006698             \\
                     & ST8000DM002   & 0.005044            & 0.007391           & 0.022137           & 0.010971          & 0.00377              \\
                     & ST8000NM0055  & 0.004534            & 0.007392           & 0.020378           & 0.020668          & 0.003742                \\ \bottomrule  
\end{tabular} 
 
\end{table*}

We first  describe the two datasets and the provided drive features.
{\noindent \textbf{SSD dataset:}} The data consists of daily performance logs for three MLC SSD models collected at a Google data center over a period of six years. All three models are  manufactured by the same vendor and have a 480GB capacity and a lithography on the order of 50nm. They utilize custom firmware and drivers, meaning that error reporting is done in a proprietary format rather than through standard SMART features~\cite{SMART}.  We refer to the three models as MLC-A, MLC-B, and MLC-D in accordance with the naming in~\cite{BiancaFAST16,mahdisoltani2017proactive}. We have data on over 10,000 unique drives for each drive model, totaling over 40,000,000 daily drive reports overall.

The logs used in this paper report daily summaries of drive activity. Drives are uniquely identified by their drive ID, which is a hashed value of their serial number. For each day of operation, the following metrics are reported:
\begin{itemize}
    \item The timestamp of the report, given in microseconds since the beginning of the drive's lifetime.
    \item The number of read, write, and erase operations performed by the drive over the course of the day.
    \item The cumulative number of P/E cycles seen by the drive over its lifetime. A program--erase cycle is the process by which a memory cell is written to and subsequently erased. The cumulative amount of these cycles is a measure of device wear.
    \item Two status flags indicating whether the drive has died and whether the drive is operating in read-only mode. 
    \item The number of {\em bad blocks} in the drive. A block is marked bad either when it is non-operational upon purchase (denoted a {\em factory bad block}) or when a non-transparent error occurs in the block and it is subsequently removed from use. Cumulative counts of both of these bad block types are provided in the log.
    \item The counts of different errors that have occurred over the course of the day. Table~\ref{table:sdd-features} lists the  error types that are logged in the dataset.
    

\begin{table*}[]
    \centering
    \caption{Name and description of different error types logged in the dataset.}
    \label{table:sdd-features}
    \begin{tabular}{|m{2.24cm}<{\centering}|m{6.15cm}<{\centering}||m{2.24cm}<{\centering}|m{6.15cm}<{\centering}|}
        \hline
        Name & Description & Name & Description \\
        \hline
         correctable error & \makecell{Number of bits that were found corrupted \\ and corrected using drive-internal error correction  \\codes  (ECC) during read operations during that day} & read error &\makecell{  Number of read operations that experienced an  error\\ but succeeded on retry (initiated drive-internally)}\\
        \hline
          erase error & \makecell{Number of erase operations that failed}  & response error & \makecell{ Number of bad responses from the drive}  \\
        \hline
          final read error & \makecell{Number of read operations that fail,\\ even after (drive-initiated) retries}   & timeout error &\makecell{Number of operations that \\ timed out after some wait period}   \\
        \hline
          final write error &\makecell{Number of write operations that fail, \\even after (drive-initiated) retries}  & uncorrectable error & \makecell{Number of uncorrectable ECC errors encountered\\ during read operations during that day}  \\
        \hline
         meta error & \makecell{Number of errors encountered while reading\\ drive-internal metadata}  & write error &\makecell{  Number of write operations that experienced an error\\ but succeeded on retry (initiated drive-internally)} \\
        \hline
    \end{tabular}
\end{table*}

\end{itemize}

{\noindent \textbf{HDD dataset:}}
The HDD dataset contains daily logs for HDDs collected at a Backblaze data center over six years. 
Basic information such as serial number, model, and date is reported, as well as standard SMART features.
Some SMART features are not reported for Hitachi drives and also for Seagate drives before 2014, for this reason we only consider Seagate drives starting from January 17, 2014, until December 31, 2019.
There are over 100,000 unique HDDs in the dataset, with more than 100,000,000 daily reports in total.
There are more than 17 different models in the dataset, of those
we consider the 5 most popular models that constitute the 
HDDs are uniquely identified by their serial number.
Every day, snapshot operation is performed for every operational hard drive, reporting the following metrics:
\begin{itemize}
    \item The timestamp of the snapshot, which is the date that the snapshot is taken.
    \item The serial number and model of the hard drive.
    \item Flag of failure, where 0 shows that the drive is functioning, while 1 means this is the last operational day before the drive fails.
    \item SMART features: All SMART features are included in the dataset, however, some of them are empty. We are interested in the non-null features, shown in Table \ref{tab:smart}.

\end{itemize}

\begin{table*}[]
    \centering
    \caption{ID, name, and description of considered SMART features.}
    \label{tab:smart}
    \begin{tabular}{|c|c|c||c|c|c|}

        \hline
        ID & Name & Description & ID & Name & Description \\
        \hline
        1 & Read Error Rate & \makecell{Frequency of errors when\\performing a read operation} & 190 & Temperature Difference & \makecell{Difference between a manufacturer-\\defined threshold (usually 100)\\and the measured temperature} \\
        \hline
        3 & Spin-Up Time & \makecell{Average time of spindle\\spin-up in milliseconds} & 192 & Power-off Retract Count & \makecell{Number of power-off\\or retract cycles} \\
        \hline
        4 & Start-Stop Count & \makecell{Number of spindle start-stop cycles} & 193 & Load/Unload Cycle Count & \makecell{Number of load/unload cycles} \\
        \hline
        5 & Reallocated Sectors Count & \makecell{Number of reallocated sectors} & 194 & Temperature & \makecell{Temperature of the hard drive} \\
        \hline
        7 & Seek Error Rate & \makecell{Frequency of errors while\\the drive head is positioning} & 197 & Current Pending Sector Count & \makecell{Number of sector\\pending for reallocation} \\
        \hline
        9 & Power-On Hours & \makecell{Number of hours of the drive\\in power-on state (drive age)} & 198 & Uncorrectable Sector Count & \makecell{Number of uncorrectable sector} \\
        \hline
        10 & Spin Retry Count & \makecell{Number of spin start retries} & 199 & UltraDMA CRC Error Count & \makecell{Number of errors identified by\\the Interface Cyclic Redundancy\\Check (ICRC) during data transfer} \\
        \hline
        12 & Power Cycle Count & \makecell{Number of cycles when the\\drive is in power-on state} & 240 & Head Flying Hours & \makecell{Number of hours when\\the drive head is positioning} \\
        \hline
        187 & Reported Uncorrectable Errors & \makecell{Number of uncorrectable\\ECC errors} & 241 & Total LBAs Written & \makecell{Number of Logical Block\\Addressing (LBAs) written} \\
        \hline
        188 & Command Timeout & \makecell{Number of aborted operations\\due to HDD timeout} & 242 & Total LBAs Read & \makecell{Number of Logical Block\\Addressing (LBAs) read} \\
        \hline
    \end{tabular}
\end{table*}

In addition to the raw features, we also consider the cumulative  effect of SMART 187 (Reported Uncorrectable Errors). 
For cumulative features (i.e., SMART 4, 5, 7, 9, 10, 12, 192, 193, 197, 198, 199, 240, 241, and 242), we also calculate their non-cumulative version (the difference with the previous observation), noted as ``diff".

\begin{table*}[tb]
\caption{Matrix of Spearman correlations among different features. We calculate the features in SSDs and HDDs separately. Bolded text indicates a large correlation value.}
    \begin{tabular}{c|lrrrrrrrrrcc}
    \toprule
\multirow{13}{*}{SSD~} &    {} &  erase &  final read &  final write &  ~~~meta & ~~~read & response & timeout &  uncorrectable &  write &  P/E cycle & bad block \\
    \cline{2-13}
&    erase             &             1.00 
    \\
 &   final read        &             0.21 &                  1.00 
    \\
  &  final write       &             0.24 &                  0.12 &                   1.00 
    \\
 &   meta              &             0.17 &                  0.19 &                   {\bf 0.35} &            1.00 
    \\
 &   read              &             0.22 &                  0.20 &                   0.30 &            {\bf 0.40} &            1.00 &                
    \\
 &   response         &             0.02 &                  0.06 &                   0.24 &            0.02 &            0.03 &                1.00 &             
    \\
 &   timeout           &             0.01 &                  0.12 &                   {\bf 0.44} &            0.02 &            0.03 &                {\bf 0.53} &               1.00 &  
    \\
 &   uncorrectable    &             0.20 &                  {\bf 0.97} &                   0.06 &            0.16 &            0.15 &                0.03 &               0.03 &                     1.00 &             
    \\
 &   write             &             {\bf 0.32} &                  0.28 &                   0.13 &            0.14 &            0.25 &                0.02 &               0.02 &                     0.28 &             1.00 &          
    \\
 &   P/E cycle            &             {\bf 0.32} &                  0.18 &                  $-0.05$ &           $-0.02$ &            0.03 &                0.03 &               0.00 &                     0.19 &             0.23 &             1.00                       
    \\
  &  bad block count ~~~&             {\bf 0.38} &                  {\bf 0.37} &                   0.19 &            0.19 &            0.18 &                0.01 &               0.01 &                     {\bf 0.37} &             {\bf 0.34} &             0.16              & 1.00           	\\
  &  drive age &  0.20 & \textbf{0.36} & 0.06 & 0.05 & 0.06 & 0.04 & 0.05 & \textbf{0.36} & 0.14 & \textbf{0.73} & 0.18 \\ \midrule
    \end{tabular}
\begin{tabular}{c|lllllllllll}
\multirow{12}{*}{HDD} &                        & \multicolumn{1}{c}{\multirow{2}{*}{failure}} & reallocated & current pending & ultraDMC & \multicolumn{1}{c}{\multirow{2}{*}{uncorrect.}} & power-off     & \multicolumn{1}{c}{\multirow{2}{*}{timeout}} & power-on       & power & \multicolumn{1}{c}{\multirow{2}{*}{temp.}} \\ 
                      &                        & \multicolumn{1}{c}{}                         & sector count     & sector count          & CRC      & \multicolumn{1}{c}{}                            & retract  cout     & \multicolumn{1}{c}{}                         & hour           & cycle & \multicolumn{1}{c}{}                             \\ \cline{2-12}
                      & failure                & 1.00                                         &             &                 &          & \textbf{}                                       &               &                                              &                &       &                                                  \\
                      & reallocated sector     & 0.03                                         & 1.00        &                 &          &                                                 &               &                                              &                &       &                                                  \\
                      & current pending sector & 0.05                                         & 0.14        & 1.00            &          &                                                 &               &                                              &                &       &                                                  \\
                      & ultraDMC CRC           & 0.01                                         & 0.02        & 0.03            & 1.00     &                                                 &               &                                              &                &       &                                                  \\
                      & uncorrectable          & 0.04                                         & 0.28        & \textbf{0.39}   & 0.07     & 1.00                                            &               &                                              &                &       &                                                  \\
                      & power-off retract count     & 0.00                                         & 0.05        & 0.00            & 0.02     & -0.02                                           & 1.00          &                                              &                &       &                                                  \\
                      & timeout                & 0.01                                         & 0.04        & 0.05            & 0.30     & 0.11                                            & 0.03          & 1.00                                         &                &       &                                                  \\
                      & power-on hours          & 0.00                                         & 0.04        & 0.04            & 0.06     & 0.08                                            & -0.21         & 0.07                                         & 1.00           &       &                                                  \\
                      & power cycle            & 0.00                                         & 0.02        & 0.04            & 0.13     & 0.08                                            & -0.07         & 0.21                                         & \textbf{0.54}  & 1.00  &                                                  \\
                      & temperature            & 0.00                                         & 0.03        & -0.01           & -0.05    & -0.03                                           & \textbf{0.33} & -0.06                                        & \textbf{-0.32} & -0.30 & 1.00             \\ \bottomrule    
\end{tabular}
\label{table:corr}
\end{table*}

{\bf Drive age and data count:}
For a given drive, the error log may have observations spanning over a period of several days up to several years. This is depicted in the ``Max Age'' CDF in Fig.~\ref{fig:agecdf}, which shows the distribution over ``oldest" observations we have for each drive. We show the CDF of SSDs and HDDs separately in Figs.~\ref{fig:agecdf}(a) and (b), respectively. This measure indicates the length of the observational horizons we possess.
We observe that, for over 50\% of drives, we have data extending over a period of 4 to 6 years for SSDs, and 2 to 5 years for HDDs, which shows that generally HDDs have a shorter lifespan.
However, days of activity are not logged for all drives. Accordingly, we may ask what magnitude of data we have access to for a given drive. The accompanying ``Data Count'' CDF shows exactly this: the number of drive days that are recorded in the error log for each drive. ``Data Count" is a measure of the volume of the log entries. Measuring them as a function of time is reasonable as there is one log entry per day per drive. 
Fig.~\ref{fig:agecdf} clearly shows that there are ample data available for both SSDs and HDDs, and therefore amenable to the  analysis presented in this paper.

Errors collected in the error logs  can be separated into two types: {\em transparent} and {\em non-transparent} errors. Transparent errors (i.e., correctable, read, write, and erase errors) may be hidden from the user while non-transparent errors (i.e., final read, final write, meta, response, timeout, and uncorrectable errors) may not.
Incidence statistics for each of these error types are listed in Table~\ref{table:different-error}.
Note that meta, response, and timeout errors in the SSD dataset are very rare. The number of uncorrectable and final read errors is at least one order of magnitude larger than other errors.
For a detailed analysis of the SSD dataset that focuses on raw bit error rates, uncorrectable bit error rates, and their relationships with workload, age, and drive wear out, we direct the interested reader to~\cite{BiancaFAST16}. 
For HDDs, only two types of errors are reported (i.e., uncorrectable error and ultraDMC CRC error). In addition to these two errors, we also consider reallocated and uncorrectable sector counts, which are highly related to errors \cite{pinheiro2007failure}, and power-off retract counts.

{\bf Correlation among errors and different features:}
Table~\ref{table:corr} illustrates the Spearman correlation matrix across pairs of all measures, aiming to determine whether there are any strong co-incidence relationships between different error types. Spearman correlations are used as a non-parametric measure of correlation, with values ranging between $-1$ and $+1$. The Spearman correlation differs from the more common Pearson correlation in that it is able to detect all sorts of monotonic relationships, not just linear ones~\cite{statistics}.
Bolded values are those with magnitude greater than or equal to $0.30$, indicating a non-negligible relationship between the pair.
We calculate the Spearman correlation matrix for the SSD and HDD datasets separately.


We first discuss the correlation of P/E cycle count with other errors in SSD dataset.
Since SSDs can only experience a finite number of write operations before their cells begin to lose their ability to hold a charge, manufacturers set a limit to the number of P/E cycles a given drive model can handle. For our drive models, this limit is 3000 cycles.
Due to these limits, it is believed that errors are caused in part by wear of write operations on the drive, which one can measure using either a cumulative P/E cycle count or a cumulative count of write operations. Using either measure is equivalent since they are very highly correlated. 

It is interesting to observe that there is little-to-no correlation of
P/E cycle count with any of the other errors, except for some moderate correlation with erase errors, which contradicts common expectations. One reason for this is the aforementioned argument regarding device wear. Another is that drives which experience more activity will simply have more opportunities for errors to occur. We are unable to detect any substantial effects even due to this na\"\i{}ve relationship.
Note that the correlation value of P/E cycle count and uncorrectable error count (which reflects bad sectors and eventual drive swap) is mostly insignificant. {
The age of a drive gives a similar metric for drive wear, which correlates highly with the P/E cycle count. The drive age also has very small correlation with cumulative error counts, with the exception of uncorrectable/final read errors.
}

Bad blocks, another likely culprit for drive swaps, shows some mild correlation with erase errors, final read errors, and uncorrectable errors. 
The high value of the correlation coefficient of 0.97 between uncorrectable errors and final read errors is not useful as the two errors represent essentially the same event: if a read fails finally, then it is uncorrectable. 
Yet, we see some moderate correlation values between certain pairs of measures that eventually show to be of use for swap prediction within the context of machine learning-based predictors, see Section~\ref{sec:prediction}.

For the HDD dataset, we present the correlation of power-on hours (which reflects drive age), power cycle, and temperature with different errors  and abnormal behaviors.
All of them have little-to-no correlation with other drive errors and  drive failures. Only temperature shows mild correlation with power-off retract count. 
Uncorrectable errors and current pending sector count have some correlation because the pending sector count reflects the number of unstable sectors in an HDD, and as more and more pending sectors appear in the drive, the more uncorrectable errors are observed.
Power cycle and power-on hour are correlated, because they are both power-related metrics.

\observations{
		\textbf{Observation \#1:} For SSDs, there is no clear relationship between non-transparent error types and uncorrectable error counts that presumably result in bad sectors and eventual drive failures.
		Program--erase (P/E) cycle counts, an indicator of drive wear, show very low correlations with most uncorrectable errors and mild correlation with erase errors (transparent errors). Drive age shows a similar pattern of correlation.
}

\observations{
		\textbf{Observation \#2:} For both datasets, correlations among all pairs of transparent and non-transparent errors show that some pairs may be mildly correlated and can be useful in prediction. Yet, there is no strong indication as to which measures are most useful for prediction.
}

\section{Drive Failure and Repair}
\label{sec:characterization}

\begin{figure}
\centering
    \subfloat[SSD]{\includegraphics[width=\columnwidth]{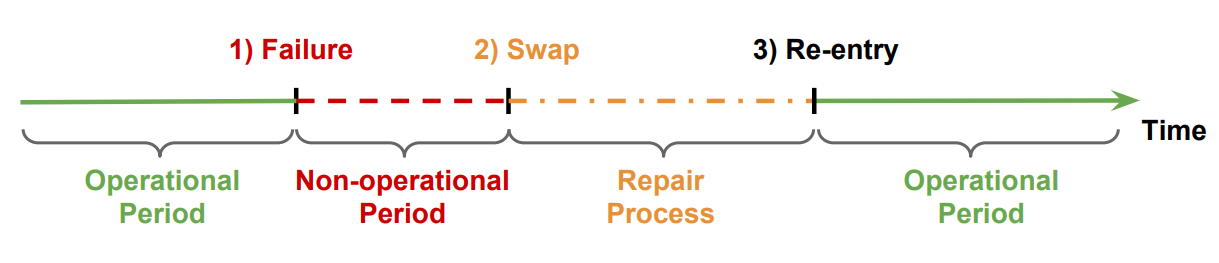}}
    
    \subfloat[HDD]{\includegraphics[width=\columnwidth]{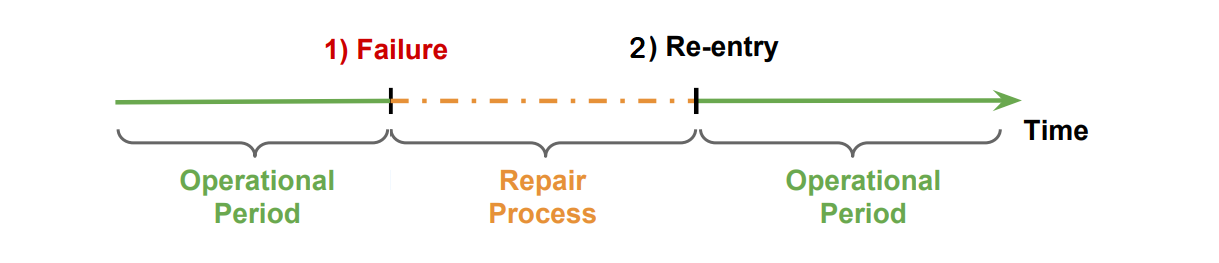}}
\caption{
Overview of failure timeline. Different from SSD, we cannot identify non-operational period and swap operation in BackBlaze dataset for HDDs.
}
\label{fig:timeline}
\end{figure}

\begin{figure}
\centering
    \begin{minipage}{0.49\columnwidth}
 		\centering
  		\includegraphics[scale=0.56]{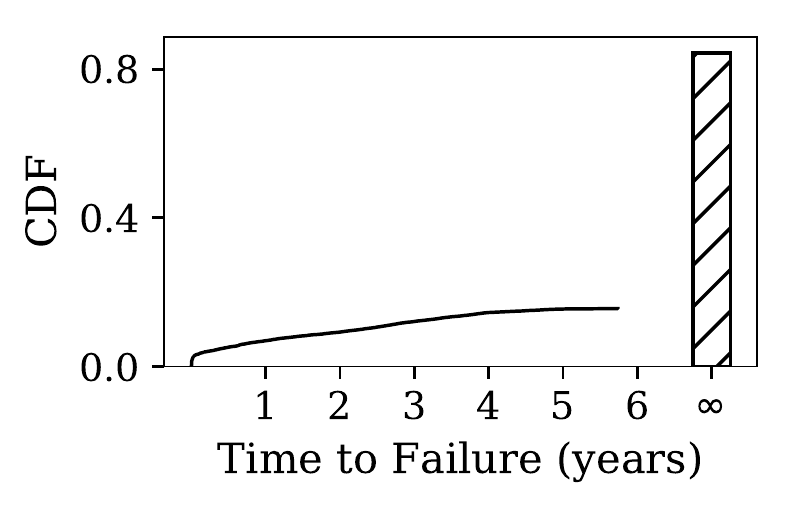}
  		
  		(a) SSD
 	\end{minipage}
 	 	\begin{minipage}{0.49\columnwidth}
 		\centering
  		\includegraphics[scale=0.56]{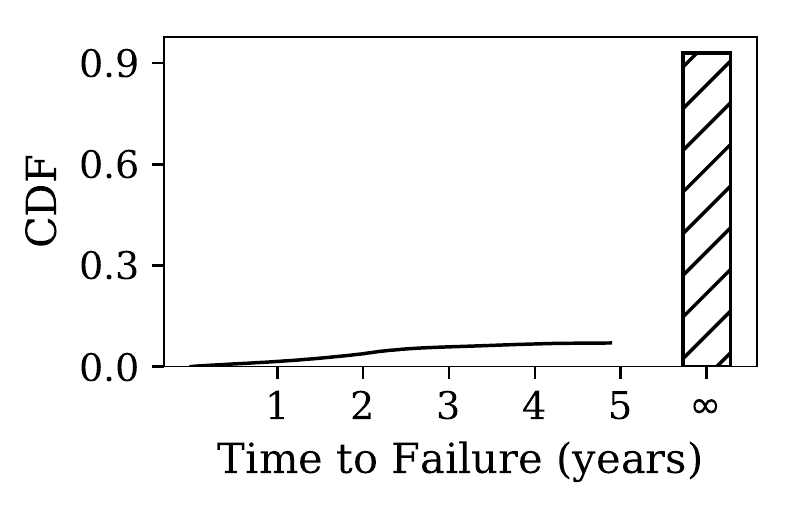}
  		
  		(b) HDD
 	\end{minipage}

\caption{
CDF of the length of the drive's operational period. The bar indicates what proportion of operational periods are not observed to end. Failure rate of SSDs is larger than HDDs.
}
\label{fig:time-to-failure}
\end{figure}

\begin{figure}
\centering
    {\includegraphics[width=0.58\columnwidth]{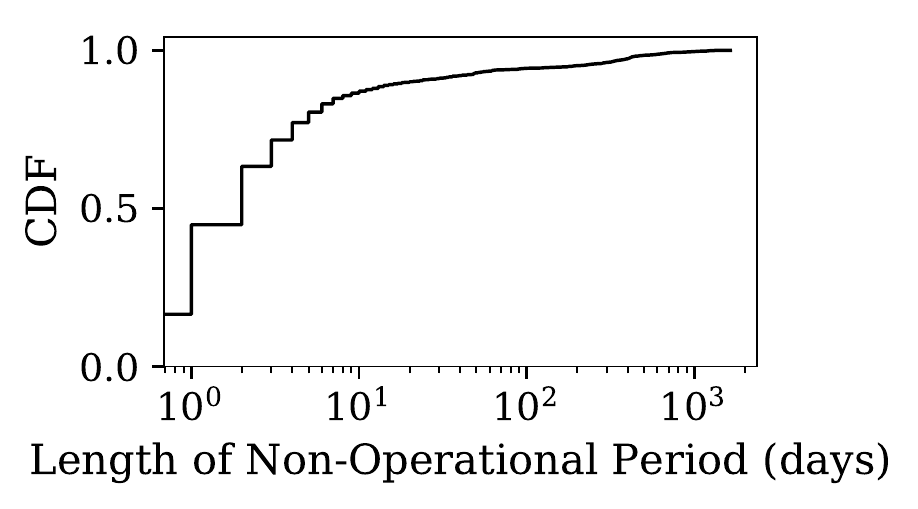}}
\caption{
CDF of length of non-operational period preceding a swap. This is the number of days between the swap-inducing failure and the physical swap itself. There is no non-operational period reported in the BackBlaze dataset for HDDs.
}
\label{fig:ssd-non-oprational-period}
\end{figure}

\begin{figure}
\centering
        \begin{minipage}{0.49\columnwidth}
 		\centering
  		\includegraphics[scale=0.56]{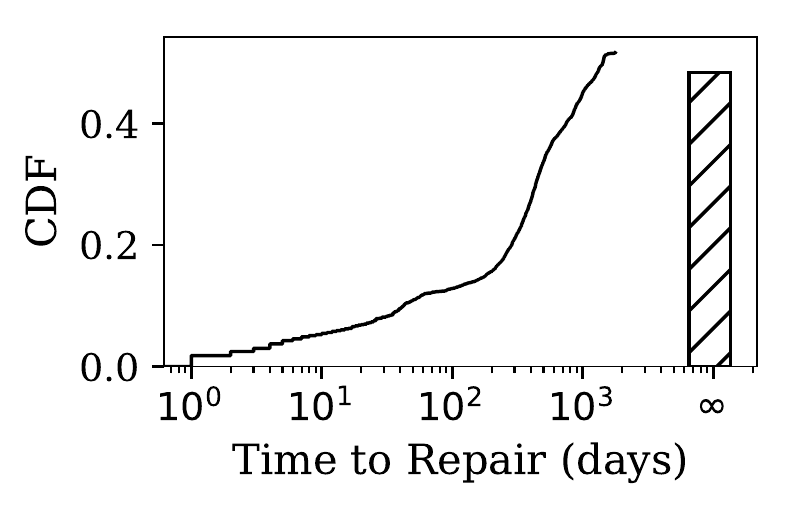}
  		
  		(a) SSD
 	\end{minipage}
 	 	\begin{minipage}{0.49\columnwidth}
 		\centering
  		\includegraphics[scale=0.56]{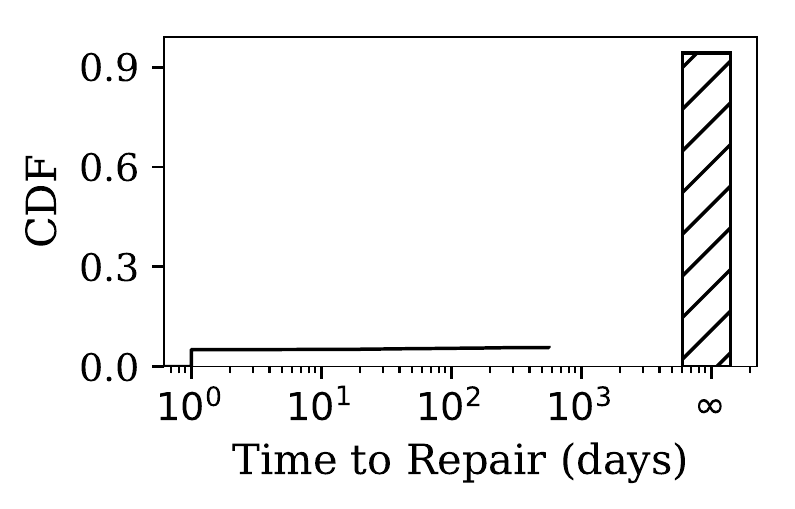}
  		
  		(b) HDD
 	\end{minipage}

\caption{
CDF of time to repair, ranging from 1 day to 4.85 years for SSD, and ranging from 1 day to 1.54 years for HDD. The proportion of repairs that are not observed to terminate is depicted with the bar graph. Half of the failed SSDs get repaired, while only 6\% of failed HDDs are repaired.
}
\label{fig:time-to-repair}
\end{figure}

In this section, we focus on drive failures and actions following a failure.
Table~\ref{table:number-failures} shows the percentage of failures for each drive model in SSD and HDD datasets. 
On average, the failure rate of SSDs is higher than the one of HDDs. 
The largest failure rate of 31.89\% is observed for ST3000DM001, an HDD model.
This high frequency of failures  poses a large pressure in terms of maintenance costs, since each failure requires manual intervention. Table~\ref{table:pct-failures-by-fail-times} provides more insights by reporting statistics on the frequency of failures for the {\em same} drive. 
Unexpectedly, we find that some SSDs have failed as many as four times over the course of their lifetime. Nonetheless 89.6\% of drives with failures, fail only once.
HDDs have no more than 2 failures per drive, and only 0.194\% of the failed drives has 2 failures, which is very rare for HDDs comparing to 9.208\% for SSDs.

\begin{table}
    \caption{High-level failure incidence statistics. This includes, for each model, the number of failures observed and the proportion of drives that are observed to fail at least once.}
    \label{table:number-failures}
	\centering
	\begin{tabular}{c|c|c|c} \toprule
	&	Model & \#Failures & \%Failed \\ \hline
\multirow{4}{*}{SSD}    &    MLC-A & 734 & 6.95 \\
    &    MLC-B & 1565 & 14.3 \\ 
    &    MLC-D & 1580 & 12.5 \\ \cline{2-4}
    &    All   & 3879 & 11.29 \\ \hline 
\multirow{6}{*}{HDD}    &    ST12000NM0007   & 1448 & 3.76 \\
& ST3000DM001   &   1357 & 31.89 \\
& ST4000DM000   &   3710 & 10.21 \\
& ST8000DM002   &    354 & 3.47 \\ 
& ST8000NM0055  &    435 & 2.91 \\ \cline{2-4}
&         All & 7743 &  7.01\\
        \bottomrule
	\end{tabular}
\end{table}

\begin{table}[htb]
	\centering
    \caption{Distribution of lifetime failure counts. The distribution is expressed with respect to the entire population of drives and with respect to those drives which fail at least once (``failed drives'').}
    \label{table:pct-failures-by-fail-times}
    \begin{tabular}{c|c|c|c}
    \toprule
  &  Number of Failures &  \% of drives &\% of failed drives \\
    \hline
      &  0 &  88.71 & ---\\
      &  1 &  10.10 & 89.60 \\
  SSD &  2 &  1.038 & 9.208 \\
      &  3 &  0.133 & 1.180 \\
      &  4 &  0.001 & 0.001 \\ \hline
      &  0 & 93.35  & --- \\
  HDD &  1 & 6.64 & 99.81\\
      &  2 & 0.01 & 0.194\\
    \bottomrule 
    \end{tabular}
\end{table}

To better characterize the conditions of failure that lead to these repairs, we must pinpoint the failures in the timeline. 

For HDDs, the  failure event is directly logged using the {\em failure} feature.
Fig.~\ref{fig:timeline}(b) shows the timeline of an HDD: 1) a failure can  happen during the operational period, then the failed HDD goes into repair process. 2) The failed drive may or may not be repaired successfully. If it is repaired, it re-enters its operational period.

For SSDs, in addition to daily performance metrics, special ``swap'' events are also reported in the data. These events indicate the time at which failed drives are extracted to be repaired.
Swaps denote visits to the repairs process -- and not simply a swapping out for storing spare parts, or moving a healthy SSD to a storage cabinet. All swaps follow drive failures, and accordingly, each swap documented in the log corresponds to a single, catastrophic failure.
A natural way to proceed is to define failure events with respect to swap events: a failure occurs on a drive's last day of {\em operational activity} prior to a swap. This is a natural point of failure since, after this point in the timeline, the drive has ceased normal function and is sent soon to the repair process.

We now discuss what we consider to be ``operational activity.''
It is often the case (roughly $80\%$ of the time) that swaps are preceded by at least one day for which {\em no performance summaries} are documented in the SSD log. This indicates that the drive was non-operational during this period, having suffered a complete failure. 
Prior to this period, we also find substantially higher rates of inactivity relatively to normal drive operation. In this case, inactivity refers to an absence of read or write operations provisioned to the drive. A period of inactivity like this is experienced prior to 36\% of swaps. The length of these inactive periods is less than one week in a large majority of cases.
The existence of such inactivity is an indication that
data center maintainers no longer provision workloads onto the drive: this amounts to a ``soft'' removal from production before the drive is physically swapped. Accordingly, we define a failure as happening directly prior to this period of inactivity, if such a period exists.

To summarize, SSD  repairs undergo the following sequence of events, represented in Fig.~\ref{fig:timeline}(a): 1) At some point, the drive undergoes a failure, directly after which the drive may cease read/write activity, cease to report performance metrics, or both, in succession.
2) Data center maintenance takes notice of the failure and swap the faulty drive with an alternate. Such swaps are notated as special events in the data. 3) After a visit to the repairs process, a drive may or may not be returned to the field to resume normal operation.

We will now characterize each of the stages, in turn, and at the same time, compare the behavior of SSDs and HDDs. We start by examining the operational periods observed in the log.
Fig.~\ref{fig:time-to-failure} presents the CDF of the length of operational periods (alternately denoted ``time to failure''). The CDF includes both operational periods starting from the beginning of the drive's lifetime and operational periods following a post-failure re-entry. It is interesting to note that more than 80\% and 90\% of operational periods of SSDs and HDDs, respectively, are not observed to end with failure during the 6 year sampling period; this probability mass is indicated by the bar centered at infinity.  
The figure indicates that there is substantial variability in the drive operational time, with the majority of operating times being long. Yet, there is a non-negligible portion of operating times that are interrupted by  failures.

Comparing SSDs with HDDs in Figs.~\ref{fig:time-to-failure}(a) and (b), the maximum time to failure of HDDs is less than 5 years, while the one of SSDs is close to 6 years.
Only 7\% of HDDs fail after an operational period, while for SSDs it is near to 20\%.
This is due to 1) the lower failure rate of HDDs compared to the one of SSDs and 2) the different maximum number of failures on a single drive (i.e., 2 for HDDs and 4 for SSDs, see Table \ref{table:pct-failures-by-fail-times}).

Fig.~\ref{fig:ssd-non-oprational-period} shows the CDF of the length of the pre-swap non-operational period (only for SSDs), i.e., the elapsed time between  the drive  failure  and  when  it  is  swapped  out  of production.
Roughly  20\%  of  failed  drives  are  removed within  a  day  and   80\%  of  failed  drives   are swapped out of the system within 7 days.
However, this distribution has a very long tail (note the logarithmic scale on the x-axis). A non-negligible proportion of failed drives (roughly 8\%) remain in a failed state up to 100 days before they are removed from production. Since these faulty drives can remain in limbo for upwards of a year, the data suggest that these drives may simply have been forgotten in the system. 

Fig.~\ref{fig:time-to-repair} depicts the CDF of the length of the repair process, (i.e., \textit{time to repair}).
For SSDs, \textit{time to repair} is the time between a swap and the following re-entry, i.e., the time of the repair process in Fig.~\ref{fig:timeline}(a).
For HDDs, \textit{time to repair} is the time of the repair process between a failure and the following re-entry as shown in Fig.~\ref{fig:timeline}(b), since swap event is not reported.
In Fig.~\ref{fig:time-to-repair}, we observe huge difference between the time to repair of SSDs and HDDs.
Half of SSDs are never observed to re-enter the field (i.e., the time to repair is infinite -- their share of probability mass is again indicated by the bar), while more than 90\% of HDDs are never successfully repaired. Among drives that \textit{are} returned to the field, the majority of SSDs remains in the repair process for upwards of a year,while most of the HDDs are repaired in less than one day.

Restating the results in Fig.~\ref{fig:time-to-repair}, Table~\ref{table:time-time-repair} illustrates the percentage of failed drives that are repaired and re-enter the system after a period of $n$ days in repair (the percentage of the successfully repaired drives as a function of all drives is also shown within  the parentheses).
These metrics show that HDDs are rarely repaired, probably due to their low cost (i.e., it is cheaper to substitute an HDD than repair it). Although roughly half of SSDs return to the production environment, the repair process may take a long time to be completed.

\observations{
        \textbf{Observation \#3:} Failed SSDs are often swapped out of production within a week, though a small portion may remain in the system  even longer than a year.
}

\observations{
		\textbf{Observation \#4:} While a significant percentage of SSDs (up to 14.3\% for MLC-B, slightly smaller percentages for other MLC types) are swapped during their lifetime, only half of failed drives are seen to successfully complete the repair process and re-enter the field.
}

\observations{
		\textbf{Observation \#5:} Of those repairs that do complete, only a small percentage of them finish within 10 days. About half of SSDs that are swapped out are not successfully repaired.
}

\observations{

		\textbf{Observation \#6:} The failure rate of HDDs is lower than SSDs, and only 7\% of failed HDDs are successfully repaired.
		
}

\begin{table*}[t]
	\centering
    \caption{Percentage of swapped drives for SSD and percentage of repaired drives for HDD that re-enter the workflow within $n$ days. The percentage of repaired drives as a function of all drives is also reported within the parentheses.}
    \label{table:time-time-repair}
    \begin{tabular}{c|l|r|r|r|r|r|r|r|r} \toprule
	&	Model & 1 day & 10 days  & 30 days  & 100 days  & 1 year & 2 years & 3 years & $\infty$  \\ \hline
\multirow{4}{*}{SSD}     &   MLC-A & 1.2 (0.09) & 3.4 (0.23) & 5.0 (0.34) & 6.1 (0.43) & 17.4 (2.61) & 37.6 (2.61) & 43.6 (3.03) & 53.4 (3.71)\\
       &  MLC-B & 2.5 (0.36) & 6.8 (0.98) & 9.4 (1.34) & 12.7 (1.81) & 25.3 (3.62) & 36.1 (5.16) & 42.7 (6.11) & 43.9 (6.28) \\ 
    &    MLC-D & 1.1 (0.14) & 4.9 (0.61) & 8.1 (1.01) & 15.8 (1.97) & 28.1 (3.51) & 43.5 (5.44) & 50.2 (6.28) & 57.6 (7.20)\\  \cline{2-10}
    & All & 1.7 (0.19) & 5.4 (0.61) & 8.0 (0.91) & 12.7 (1.43) & 25.0 (2.82) & 39.4 (4.45) & 45.9 (5.19) & 51.3 (5.79)   \\\hline
 \multirow{6}{*}{HDD} &   ST3000DM001 & 24.98 (7.97)  & 25.28 (8.06)  & 25.42 (8.11)  & 25.42 (8.11)  & 25.42 (8.11)  & 25.42 (8.11)  & 25.42 (8.11)  & 74.58 (23.78) \\
& ST4000DM000 & 0.62 (0.06)   & 0.62 (0.06)   & 0.7 (0.07)    & 0.81 (0.08)   & 0.97 (0.1)    & 0.97 (0.1)    & 0.97 (0.1)    & 99.03 (10.11) \\
& ST8000NM0055 & 0.0 (0.0)    & 0.0 (0.0)     & 0.23 (0.01)   & 0.69 (0.02)   & 1.15 (0.03)   & 1.15 (0.03)   & 1.15 (0.03)   & 98.85 (2.87)  \\
& ST8000DM002 & 0.28 (0.01)   & 0.28 (0.01)   & 0.85 (0.03)   & 1.69 (0.06)   & 2.54 (0.09)   & 2.82 (0.1)    & 2.82 (0.1)    & 97.18 (3.37) \\
& ST12000NM0007 & 0.48 (0.02) & 0.55 (0.02)   & 0.76 (0.03)   & 1.04 (0.04)   & 1.45 (0.05)   & 1.45 (0.05)   & 1.45 (0.05)   & 98.55 (3.7) \\ \cline{2-10}
 &        All & 4.83 (0.32) & 4.9 (0.33) & 5.05 (0.34) & 5.29 (0.35) & 6.17 (0.41) & 6.21 (0.42) & 6.21 (0.42) & 93.79 (6.28) \\

        \bottomrule
	\end{tabular}
\end{table*}

\section{Symptoms and Causes of Failures}
\label{sec:causes}



    


In this section we make an effort to connect the statistics from the two logs, aiming to identify causes of drive failures. 
Since the features/information provided by the two datasets are different, here we analyze SSDs and HDDs separately.

\subsection{Age and Device Wear}
\label{subsec:age}

\begin{figure}
\centering
           \begin{minipage}{0.49\columnwidth}
 		\centering
  		\includegraphics[scale=0.45]{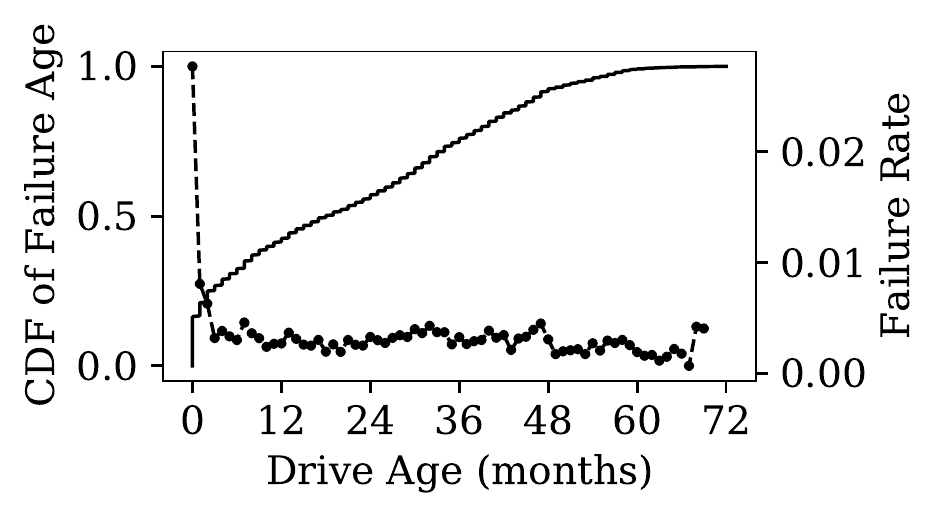}
  		
  		(a) SSD
 	\end{minipage}
 	 	\begin{minipage}{0.49\columnwidth}
 		\centering
  		\includegraphics[scale=0.45]{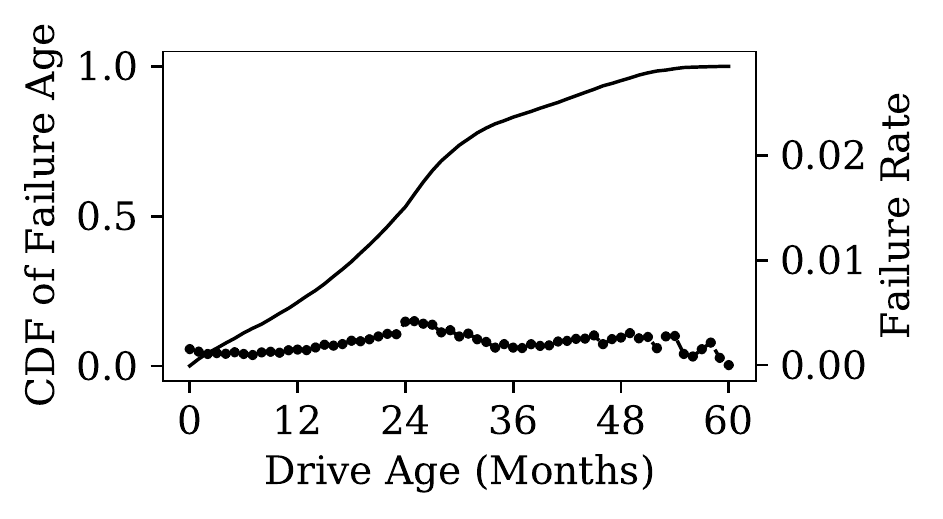}
  		
  		(b) HDD
 	\end{minipage}

\caption{
The CDF of the age of failed drives (solid line) and the proportion of functioning drives that fail at a given age level, in months (dashed line). For SSDs, this guides us to consider young and old SSDs separately, such separation is not suggested for HDDs.
}
\label{fig:age}
\end{figure}

Recall that Table~\ref{table:number-failures} shows that, among the drives represented in the datasets, 11.29\% of SSDs and 7.01\% of HDDs fail at least once.
A natural question is when do these failures (and the following swaps for SSDs) occur in the drive's lifetime: what is the role of age in drive failure?
Figure~\ref{fig:age}(a) reports the CDF of the failure age (solid line) as a function of the drive age for SSDs. The figure shows that there are many more drive failures in the first 90 days of drive operation than at any other point in the drive lifetime. In fact, 15\% of observed failures occur on drives less than 30 days old and 25\% occur on drives less than 90 days old. This seems to indicate that these drives have an infancy period during which drive mortality rate is particularly high. This performance pattern has been noticed previously in similar studies of SSDs in the wild~\cite{MezaWKM15}.

The slope of the CDF in Figure~\ref{fig:age}(a) gives us an estimate of the rate at which swaps occur at a given drive age. However, this estimate is skewed since not all drive ages are equally represented in the data. For example, the rate of failures seems to slow down following the four year mark, but this is due to the fact that drives of this age level are not as common in the data. We hence normalize the number of swaps within a month by the amount of drives represented in the data at that month to produce an unbiased failure {\em rate} for each month (dashed line in Figure~\ref{fig:age}). We see that this rate evens out after the third month, indicating that the length of this observed high-failure infancy period is approximately 90 days. Accordingly, for the remainder of this paper, we distinguish drive swaps as {\em young} versus {\em old}, i.e., those swaps occurring before vs.~after the 90-day mark. Beyond the 90-day mark, we observe that the failure rate is roughly constant, suggesting that, even if drives become very old, they are not more prone to failure.

\begin{figure}
    \centering
    \includegraphics[width=\linewidth]{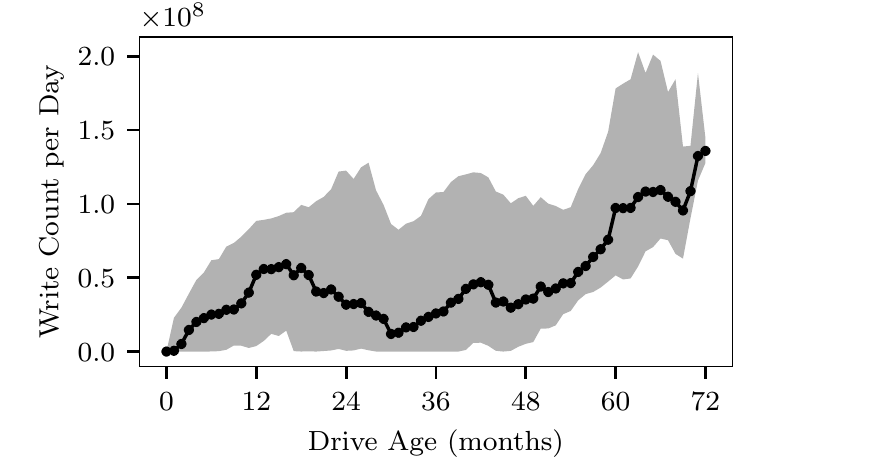}
    \caption{Quartiles of the daily write intensity per month of drive age. The line plot shows the median write intensity for each month. The 1st and 3rd quartiles are shown as the boundaries of the shaded area.}
    \label{fig:writerates}
\end{figure}

One potential explanation for the spike in failures for infant drives is that they are undergoing a ``burn-in'' period. This is a common practice in data centers, wherein new drives are subjected to a series of high-intensity workloads in order to test their resilience and check for manufacturing faults. These increased workloads could stress the drive, leading to a heightened rate of failure. To test this hypothesis, we looked at the intensity of workloads over time. For each month of drive age, we examined drives of that age and how many write operations they processed per day. The distributions of these write intensities are presented in Figure~\ref{fig:writerates}.

It is clear that younger drives do \textit{not} tend to experience more write activity than usual (in fact, they tend to experience markedly \textit{fewer} writes!). A similar trend is apparent for read activity (not pictured). We conclude that there is no burn-in period for these drives and that the spike in failure rates is caused by manufacturing malfunctions not caught by drive testing.

\begin{figure}
 \centering
    \includegraphics[width=\columnwidth]{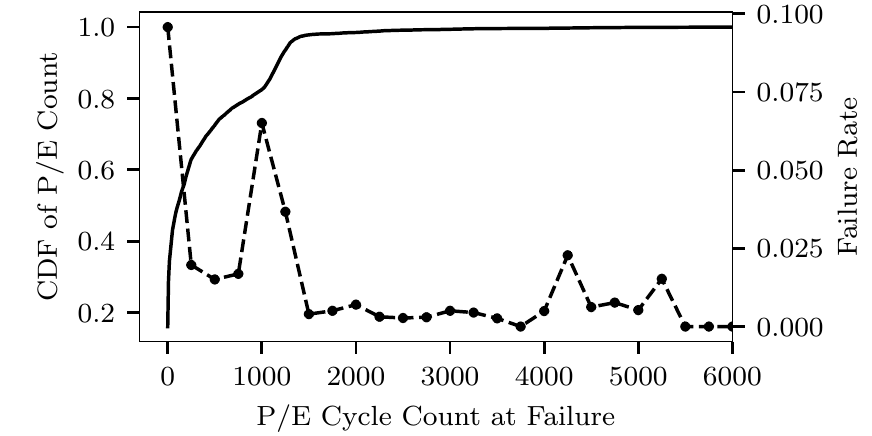}
    \caption{The distribution  P/E cycle counts of failed drives (solid) and the proportion of drives that fail (dashed) at a given P/E level, binned in increments of 250 cycles. }
    \label{fig:swapPEpdf} 
\end{figure}

Beyond drive age, we are also interested in the relationship between failure and device wear, which we measure using P/E cycles, as discussed in Section~\ref{sec:dataset}. In the same style as Figure~\ref{fig:age}, Figure~\ref{fig:swapPEpdf} illustrates 
the relationship of cumulative P/E cycles and probability of failure in the form of a CDF (solid line) and an accompanying failure rate (dashed line).
The CDF illustrates that almost 98\% of failures occur before the drive sees 1500 P/E cycles. This is surprising, considering that the manufacturer guarantees satisfactory drive performance up until 3000 P/E cycles. Conversely, the failure rate beyond the P/E cycle limit is very small and roughly constant. The spikes at 4250 and 5250 P/E cycles are artifactual noise attributed to the fact that the number of drives that fail at these P/E levels are so few in number.

\begin{figure}
    \centering
    \includegraphics[width=\columnwidth]{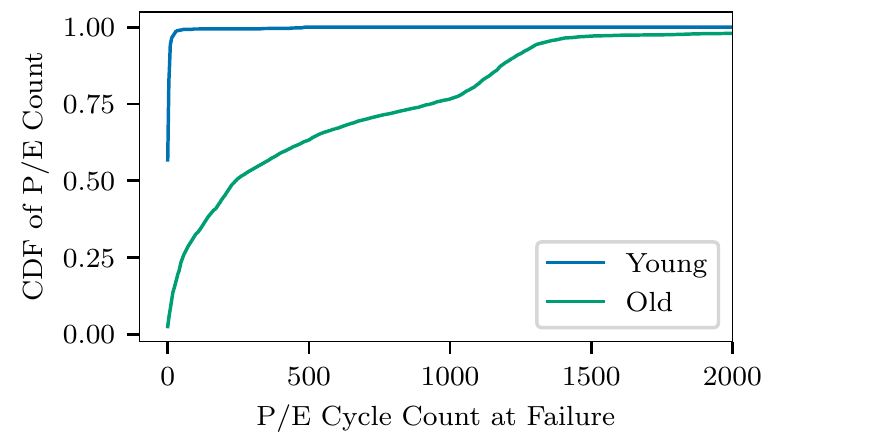}
    \caption{The CDF in Figure~\ref{fig:swapPEpdf} split across infant failures (occurring at age $\le 90$ days) and mature failures (occurring at age $> 90$ days).}
    \label{fig:pecountsplit}
\end{figure}

In the figures discussed, we observe high failures rates for both SSDs younger than three months and SSDs with fewer than 250 P/E cycles. Due to the correlation between age and P/E cycles, these two characterizations may be roughly equivalent, describing the same phenomenon. However, we do not find this to be the case. To illustrate this, we plot two CDFs in Figure~\ref{fig:pecountsplit}: one for young failures and one for old ones. It is clear that the young failures inhabit a distinct, small range of the P/E cycle distribution. Since this range is so small, the individual P/E cycle counts are not informative to young failures.

\observations{
		\textbf{Observation \#7:} Age plays a crucial role in the SSD failure/swap incidence. In particular, drives younger than 90 days have markedly higher failure incidence rates. {This phenomenon is characteristic to young drives and cannot be explained with P/E cycle counts.}
}

\observations{
		\textbf{Observation \#8:} Beyond the infancy period, age does not seem to have an important part to play in failure rate. The oldest drives seem to fail with roughly the same frequency as young, non-infant drives.
}

\observations{
		\textbf{Observation \#9:} The vast majority of drive failures happen well before the P/E cycle limit is reached. Furthermore, drives operating beyond their P/E cycle limit have very low rates of failure.
}

\subsection{Head Flying Hours in HDDs}
\label{subsec:hdd-smart-240}

Similar to the drive age in SSDs shown in Figure~\ref{fig:age}(a), we also present a similar plot for HDDs in Figure~\ref{fig:age}(b).
Differently from the high failure rate of young SSDs, the failure rate for HDDs regarding drive age is relatively small (less than 1\%). 
Therefore, we need to find other features which may be related to failure rate.

We examine {\em all SMART features} for HDDs, and find out that {\em head flying hours (HFH, SMART 240)} is highly related to failures.
Here we define two kinds of disks regarding head flying hours with a certain $threshold$: 
1) {\em Large HFH} disks are observed at least once with head flying hours larger than the $threshold$;
2) {\em Small HFH} disks always have head flying hours smaller than or equal to the $threshold$.

\begin{figure}
\centering
              \begin{minipage}{0.49\columnwidth}
 		\centering
  		\includegraphics[scale=0.5]{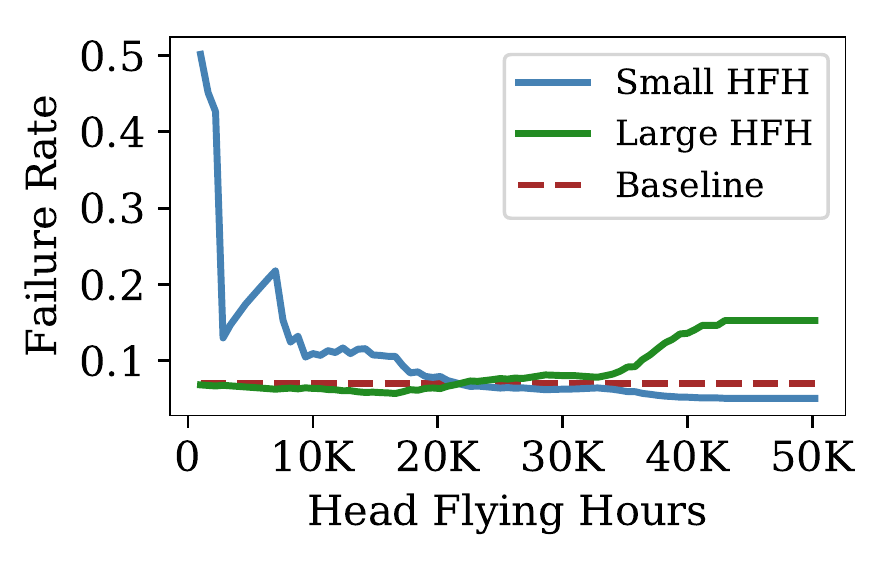}
  		
  		(a) Failure Rate
 	\end{minipage}
 	 	\begin{minipage}{0.49\columnwidth}
 		\centering
  		\includegraphics[scale=0.5]{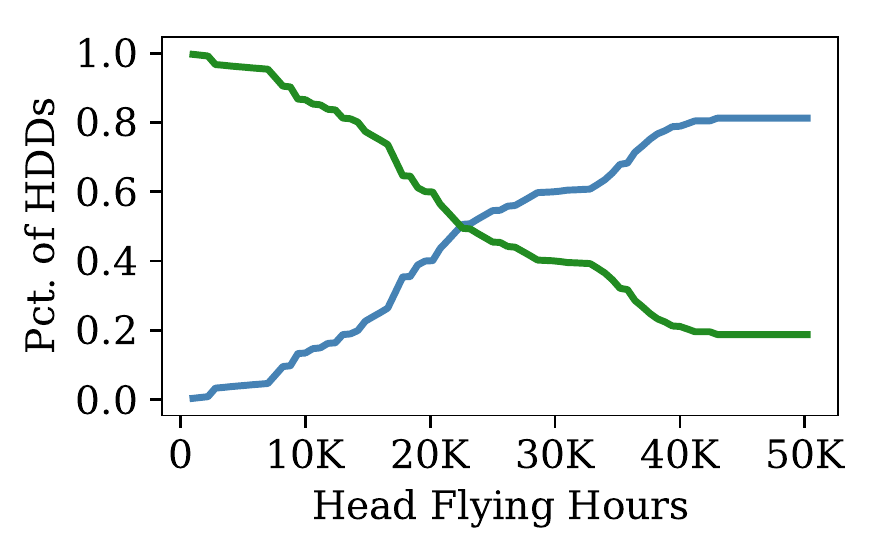}
  		
  		(b) HDD Percentage 
 	\end{minipage}

\caption{
head flying hours (SMART 240). Many HDDs fail when they have a large head flying hours. The dashed-line in subfigure (a) marks the mean failure rate.
}
\label{fig:head-flying-hours}
\end{figure}

Figure~\ref{fig:head-flying-hours}(a) shows the failure rate of small and large HFH drives as a function of the threshold.
The {\em average} faialure rate of all HDDs is also reported (baseline, see dashed line).
We observe two situations with high failure rate in Figure~\ref{fig:head-flying-hours}: 1) small HFH when threshold is less than 3000 (the beginning of small HFH line), and 2) large HFH when threshold is larger than 40,000 (the end of large HFH line).
The percentage of HDDs if we partition the dataset according to HFH is also shown in Figure~\ref{fig:head-flying-hours}(b). 
When the threshold is less than 3000, the percentage of small HFH drives is less than 3\%, therefore it is not representative.
When the threshold is larger than 40,000, the percentage of large HFH is about 20\%, which is worth to be considered.
The failure rate of these 20\% large HFH disks is 17\%, which is much higher than the 7\% failure average (baseline).
Balancing the failure rate and percentage of large HFH disks, we  use  as $threshold=40,000$ to split the dataset.
This observation (small and large have different resilience behavior) guides us to split the dataset for better prediction, see Section~\ref{subsec:improvements} for more details.

\observations{
		\textbf{Observation \#10:} 
		HDDs with large head flying hours (SMART 240) have more uncorrectable errors before failure events comparing to drives with small head flying hours.
}

\subsection{Error Incidence for SSDs}

\begin{figure}
	\centering
    
                  \begin{minipage}{0.49\columnwidth}
 		\centering
  		\includegraphics[scale=0.48]{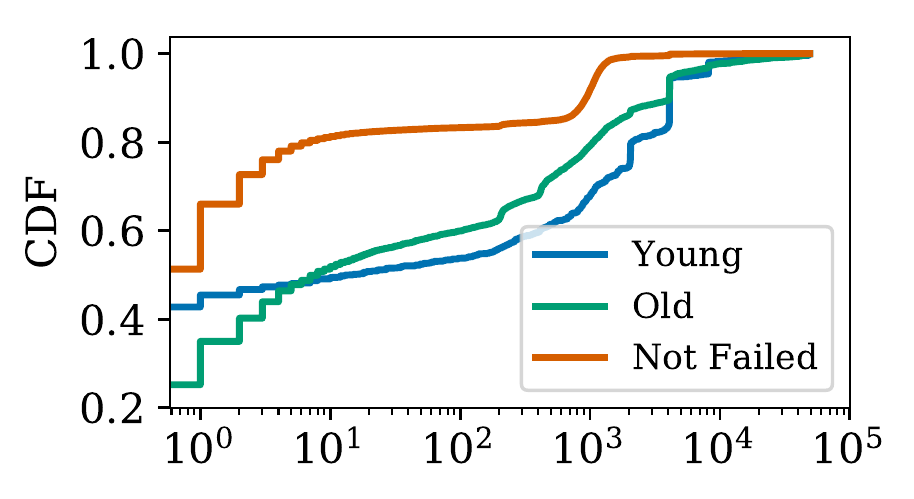}
  		
  		(a) Cumulative Bad Block Count
 	\end{minipage}
 	 	\begin{minipage}{0.49\columnwidth}
 		\centering
  		\includegraphics[scale=0.48]{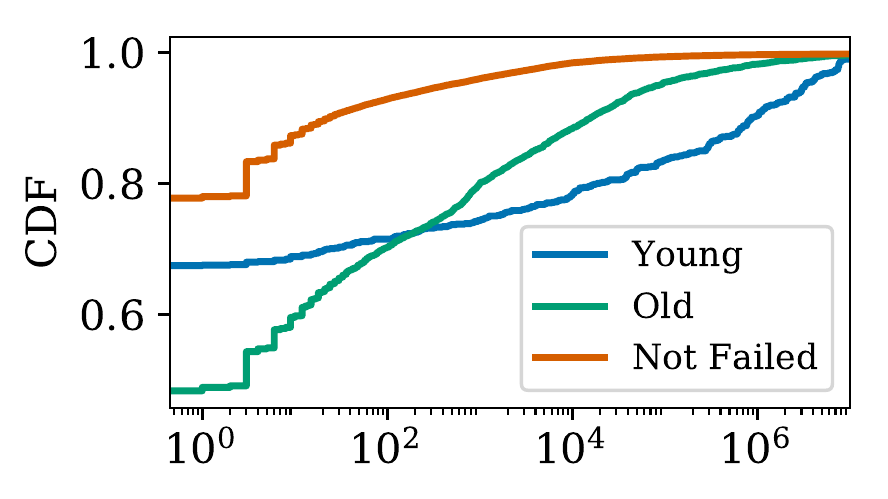}
  		
  		(b) Cumulative Uncorrectable Error Count
 	\end{minipage}

    \caption{CDF of cumulative bad block counts and uncorrectable error counts of SSDs based on the age at which the swap occurred. The ``Not Failed" CDF corresponds to the distribution over drives that are not observed to fail. }
    \label{fig:cfrecdf}
\end{figure}

Intuitively, we would expect that catastrophic drive failure is preceded by previous lapses in drive function, indicated in our data as non-transparent errors. We focus on uncorrectable errors and bad blocks since they are by far the most common of these errors. Other errors occur far too rarely to give much insight. We test the validity of our intuition by comparing the cumulative counts of errors seen by failed drives to a baseline of cumulative error counts taken across drives that are not observed to experience failure. We are also particularly interested to see if there is any difference in error incidence between young failures ($\le 90$ days) and old failures ($> 90$ days). This is illustrated with CDFs in Figure~\ref{fig:cfrecdf}. 

We find that drives that fail tend to have experienced orders of magnitude more uncorrectable errors and bad blocks than we would expect, on average. This is exemplified by the fact that in roughly 80\% of cases, non-failed drives are not observed to have experienced any uncorrectable errors. On the other hand, for failed drives, this proportion is substantially lower: 68\% for young failures and 45\% for old drives. In fact, broadening our scope, we find that 26\% of failures happen to drives which have experienced no non-transparent errors \textit{and} which have developed no bad blocks. Furthermore, we find that, if errors \textit{are} observed, then young failures tend to see more of them than old failures. This is most easily seen in the tail behavior of the aforementioned CDFs; for example, the 90th percentile of the uncorrectable error count distribution is two orders of magnitude larger for young failures than for old failures, in spite of the fact that the young drives have been in operation for much less time.

Overall, the presence of errors is not a very good indicator for drive failure since most failures occur without having seen \textit{any} uncorrectable errors. However, drives that experience failure do have a higher rate of error incidence, which means that we expect error statistics to be of some utility in failure prediction (to be discussed in Section~\ref{sec:prediction}). Furthermore, we find that the patterns of error incidence are markedly different among young and old failures. In particular, young failures have a predilection toward extremely high error counts.

\begin{figure}
    \centering
    	 	\begin{minipage}{0.49\columnwidth}
 		\centering
  		\includegraphics[scale=0.51]{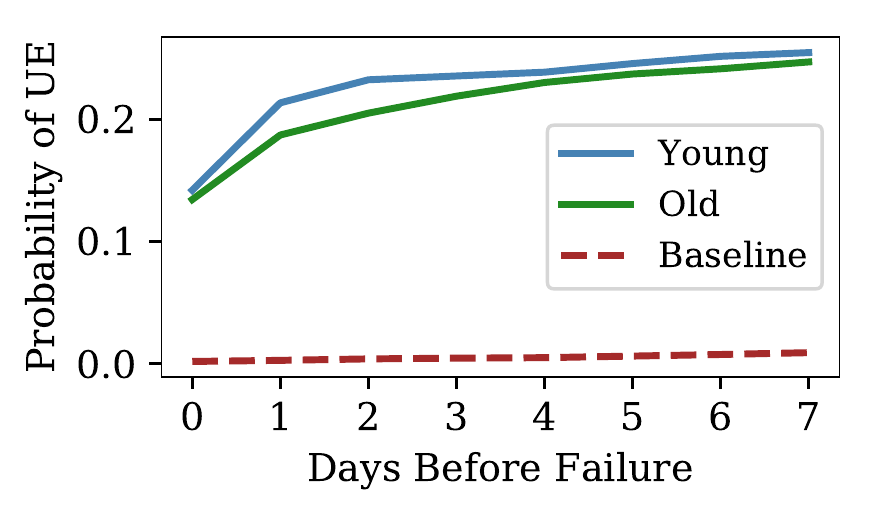}
  		
  		(a) SSD: UE probability
 	\end{minipage}
 	 	\begin{minipage}{0.49\columnwidth}
 		\centering
  		\includegraphics[scale=0.45]{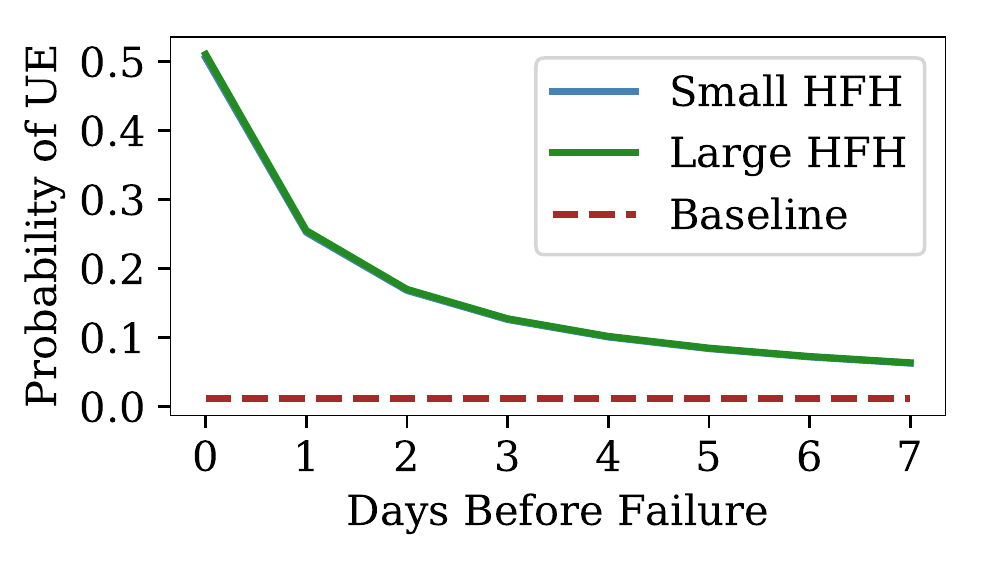}
  		
  		(b) HDD: UE probability
 	\end{minipage}
  
      \centering
    	 	\begin{minipage}{0.49\columnwidth}
 		\centering
  		\includegraphics[scale=0.5]{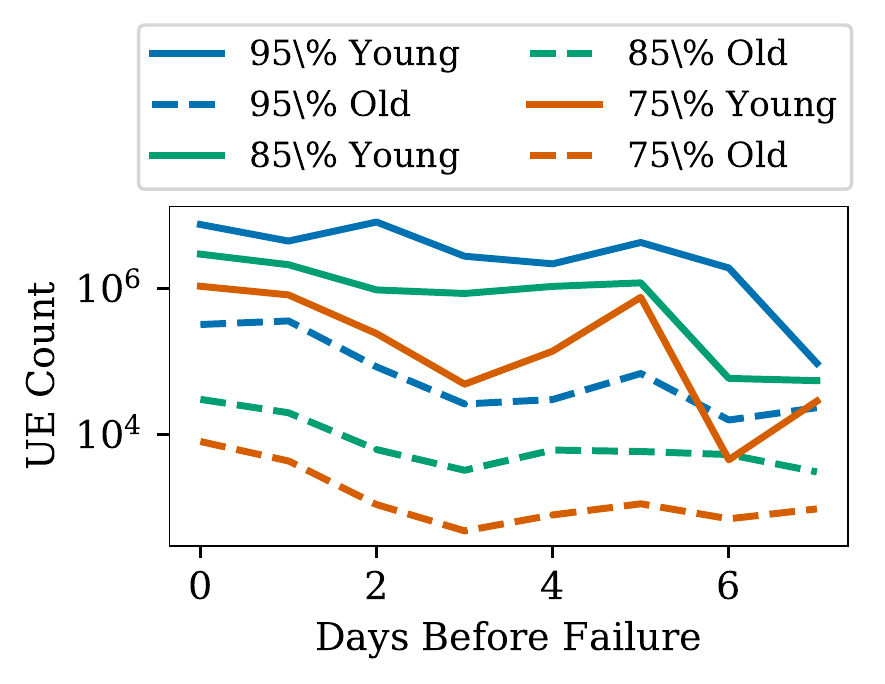}
  		
  		(c) SSD: UE count
 	\end{minipage}
 	 	\begin{minipage}{0.49\columnwidth}
 		\centering
  		\includegraphics[scale=0.45]{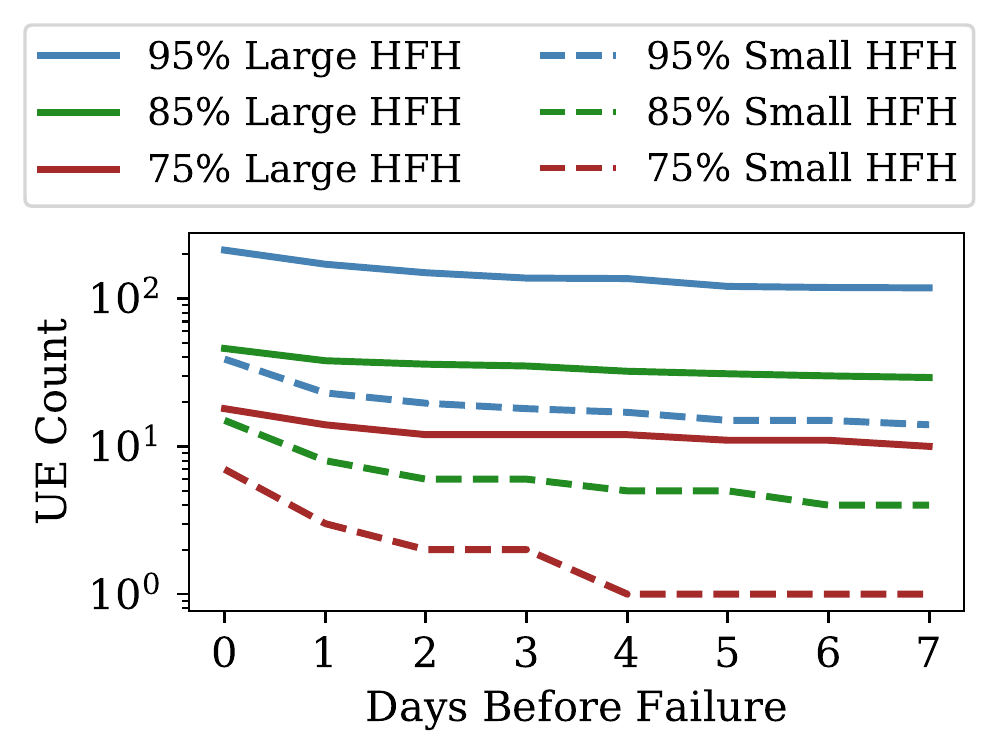}
  		
  		(d) HDD: UE count
 	\end{minipage}

    \caption{(Top) Probability of uncorrectable error (UE) happening within the last $n$ days before a swap. The baseline curve on the top graph is the probability of seeing an uncorrectable error within an arbitrary $n$-day period. (Bottom) Provided a UE happens, how many occur? Upper percentiles of the distribution of uncorrectable error counts preceding to failure, excluding zero counts.
    }
    \label{fig:uncorrectable-errors}
\end{figure}

Moving into a finer temporal granularity, we are interested in error incidence directly preceding the failure. This behavior is of particular importance for failure forecasting and prediction. We ask: do drives tend to be more error-prone right before a failure occurs? How long before the drive failure  is this behavior noticeable? Figure~\ref{fig:uncorrectable-errors} shows two relevant uncorrectable error statistics in the period before a drive swap.

Figure~\ref{fig:uncorrectable-errors}(a) shows the probability that a faulty drive had an error within the last $N$ days before its failure. The baseline is the probability of seeing an uncorrectable error within an arbitrary $N$-day period. We see that failed drives see uncorrectable errors with a much higher than average probability and that this behavior is most noticeable in the last two days preceding the failure. 
However, the probability that a failed drive does not see any errors in its last 7 days is very high (about 75\%).

Figure~\ref{fig:uncorrectable-errors}(c) shows the distribution of those uncorrectable error counts that are nonzero on each day preceding the swap. We find that error counts tend to increase as the failure approaches. We also find that young failed drives, if they suffer an error, tend to experience orders of magnitude more errors than older ones, note the log-scale on the y-axis of the graph.

To summarize, we zoom in specifically on the period directly preceding a failure. We show that error incidence rates depend on two factors: (1) the age of the drive (young vs.~mature) and (2) the amount of time until the swap occurs. The resulting increase in error rate is most noticeable in the two days preceding the swap, suggesting that the lookahead window within which we can accurately forecast failure may be small.

\observations{
	\textbf{Observation \#11:} Incidence of non-transparent errors is not strongly predictive of catastrophic drive failure. In fact, a substantial proportion of drives experience failure without having ever seen any sort of serious soft error.
}

\observations{
	\textbf{Observation \#12:} Failures of young drives are more likely to have seen higher rates of error incidence than failures of mature drives.
}

\observations{
	\textbf{Observation \#13:} Error incidence rates increase dramatically in the two days preceding a drive failure.
}

\subsection{Error Incidence for HDDs}

We also show the percentage and count of uncorrectable errors (SMART 187) for HDDs, in Figure~\ref{fig:uncorrectable-errors}(b) and (d).
For HDDs, the probability of uncorrectable errors increases when a failure approaches. 
When a failure occurs, the probability of uncorrectable errors is around 50\%. 
The probability of uncorrectable errors of small HFH and large HFH is essentially the same.
For both cases, the probability is much higher than the baseline (probability of error).

When looking into the  number of uncorrectable errors, the uncorrectable error count of HDDs is several orders of  magnitude lower than SSDs.
For HDDs, we also observe  higher uncorrectable error count for large HFH disks before failure events.

\observations{
	\textbf{Observation \#14:} Error incidence rates becomes extremely high in the two days before an HDD fails. 
}

\section{Failure Prediction}
\label{sec:prediction}

In this section, we use machine learning models to detect SSD and HDD failures that will occur within $N \geq 0$ days.
Symptoms and causes of SSD and HDD failure investigated in Section \ref{sec:causes} are used to improve model accuracy.
Identifying
feature importance allows identifying those attributes that are more critical for SSD and HDD lifetimes.

\subsection{Model description}
\label{subsec:model}
\noindent \textbf{Input.} 
Both  SSD and HDD datasets present daily statistics and are extremely imbalanced, i.e., the number of healthy disks (majority class) is larger than the number of faulty ones (minority class).
In the SSD case, the ratio of healthy and defective drives is 10,000:1, for HDDs it is 13,000:1.
To deal with such imbalanced datasets, we under-sample the majority class, use cross-validation for training and testing the model, and evaluate its performance with measures that are not affected by imbalanced datasets.

\noindent \textbf{Under-sampling.} 
For both datasets, the training set is under-sampled to result in a 1:1 healthy-faulty drives ratio to avoid the classifier being biased toward \textit{healthy} drives.
For this purpose, we use a random strategy, i.e., observations to be removed are randomly chosen from the majority class.
We observe that the model performance is not profoundly affected by considering different under-sampling strategies and healthy-faulty ratios.

\noindent \textbf{Cross-Validation.} Classifiers are cross-validated by splitting each dataset into five different folds.
The dataset is partitioned by drive ID (i.e., all observations of a drive belong to the same fold and are not used concurrently for training and testing).
If folds are created by randomly partitioning the dataset, i.e., the strategy adopted in \cite{botezatu2016predicting,mahdisoltani2017proactive}, it is possible that future observations of a drive are used to predict the past failure state of the same drive.
This is undesirable since no future information is available in real scenarios.
This is also avoided by using online prediction \cite{XuATC18}, i.e., the model is trained on observations collected before a specific date and tested on the remaining data. Unfortunately, this cannot  be applied to the SSD dataset: its observations do not have a global timestamp attribute that would allow synchronizing the various traces across time.
Four out of five folds are used for training, while the remaining one is used for testing.
Five different classifiers are trained (each one tested on a different testing set).
Global performance is obtained by averaging the performance of each classifier.

\noindent \textbf{Output.} The model returns a continuous value in the interval $(0,1)$, i.e., the probability that the drive fails.
In real-world scenarios, a binary output (i.e., failure vs. non-failure) is preferred.
For this purpose, we set a discrimination threshold, $\alpha$, that discretizes the returned probability: if the output is larger than $\alpha$, then the model predicts a failure; otherwise, the model predicts a non-failure.
Due to its insensitiveness to imbalanced datasets, receiver operating characteristic (ROC) is generally used to evaluate the accuracy of binary classifiers \cite{fawcett2006introduction,mahdisoltani2017proactive} and is adopted also in this paper.
ROC curve plots the true positive rate (TPR or recall) against the false positive rate (FPR) of the analyzed classifier by considering different values of $\alpha$.
TPR and FPR are defined as:
\begin{equation*}
TPR = \frac{TP}{TP+FN} \quad \text{and} \quad FPR = \frac{FP}{FP+TN},
\end{equation*}
where TP is the number of true positives, TN is the number of true negatives, FP is the number of false positives, and FN is the number of false negatives.
We also consider the area under the ROC curve (i.e., AUROC) as a further measure to determine the goodness of the proposed classifier.
The AUROC is always in the interval $(0.5,1)$: it is 0.5 if the prediction is not better than the one of a random classifier; its value is 1 for a perfect predictor.

\subsection{Prediction Accuracy}
\label{subsec:rf}
To determine which classifier must be used, we investigate and report the performance of different classification models in Table \ref{tab:predictors_auc}.
It shows the AUROC of predictors for different lookahead windows and datasets (i.e., SSDs and HDDs).
Although Table \ref{tab:predictors_auc} shows only one value for each classifier, we investigate their performance with different hyperparameters (e.g., the number of estimators for the Random Forest, the number of neighbors for the k-NN, or the maximum depth for XGBoost).
As observed in \cite{alter2019ssd} and \cite{mahdisoltani2017proactive} for SSDs and HDDs, respectively, Random Forest performs better than other predictors (except XGBoost) when applied to these datasets.
That may be due to the ability of Random Forests to model non-linear effects and provide high accuracy even when trained with a few points.
Although the performance of XGBoost and Random Forest models are similar (in fact, XGBoost models may be slightly more accurate than Random Forest ones), the time required to train the latter classifier is only 5\% of the time required for training the former one.
Table \ref{tab:predictors_auc} also shows that, independently of the considered classifier, the largest AUROC is obtained when the lookahead window is smaller.
\begin{table*}[t]
    \centering
    \caption{AUROC for different predictors and lookahead windows, $N$. The cross-validated average AUROC is provided with the standard deviation across folds.}
    \label{tab:predictors_auc}
    \begin{tabular}{c|c|cccc}
        \hline
        & N (lookahead days) & 0 & 1 & 2 & 7 \\
        \hline
        \multirow{6}{*}{SSD} & Logistic Reg. & $0.796 \pm 0.010$ & $0.765 \pm 0.009$ & $0.745 \pm 0.007$ & $0.713 \pm 0.010$ \\
        & k-NN & $0.816 \pm 0.013$ & $0.791 \pm 0.009$ & $0.772 \pm 0.008$ & $0.716 \pm 0.008$ \\
        & SVM & $0.821 \pm 0.014$ & $0.795 \pm 0.011$ & $0.778 \pm 0.011$ & $0.728 \pm 0.011$ \\
        & Neural Network & $0.857 \pm 0.007$ & $0.828 \pm 0.004$ & $0.803 \pm 0.009$ & $0.770 \pm 0.008$ \\
        & Decision Tree & $0.872 \pm 0.007$ & $0.840 \pm 0.007$ & $0.819 \pm 0.005$ & $0.780 \pm 0.006$ \\
        & XGBoost & $0.904 \pm 0.001$ & $0.873 \pm 0.002$ & $0.851 \pm 0.001$ & $0.809 \pm 0.001$ \\
        & Random Forest & $0.905 \pm 0.008$ & $0.859 \pm 0.007$ & $0.839 \pm 0.006$ & $0.803 \pm 0.008$ \\
        \hline
        \multirow{6}{*}{HDD} & Logistic Reg. & $0.668 \pm 0.001$ & $0.669 \pm 0.001$ & $0.668 \pm 0.001$ & $0.668 \pm 0.001$ \\
        & k-NN & $0.699 \pm 0.022$ & $0.699 \pm 0.026$ & $0.701 \pm 0.029$ & $0.691 \pm 0.029$ \\
        & SVM & $0.679 \pm 0.012$ & $0.689 \pm 0.009$ & $0.685 \pm 0.011$ & $0.684 \pm 0.009$ \\
        & Neural Network & $0.684 \pm 0.065$ & $0.683 \pm 0.076$ & $0.685 \pm 0.077$ & $0.682 \pm 0.076$ \\
        & Decision Tree & $0.886 \pm 0.051$ & $0.870 \pm 0.053$ & $0.862 \pm 0.053$ & $0.837 \pm 0.052$ \\
        & XGBoost & $0.904 \pm 0.001$ & $0.888 \pm 0.001$ & $0.878 \pm 0.001$ & $0.854 \pm 0.001$ \\
        & Random Forest & $0.903 \pm 0.013$ & $0.888 \pm 0.010$ & $0.878 \pm 0.008$ & $0.854 \pm 0.006$ \\
        \hline
    \end{tabular}
\end{table*}

\begin{figure}
    \centering
    \includegraphics[width=\columnwidth]{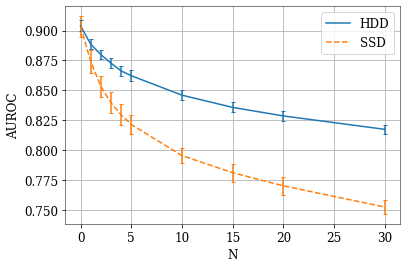}
    \caption{Random Forest AUROC as a function of lookahead window size, $N$. Error bars indicate the standard deviation of the cross-validated error across folds.}
    \label{fig:rf_auc}
\end{figure}

Figure \ref{fig:rf_auc} plots the AUROC of the Random Forest prediction on HDD (solid line) and SSD (dashed line) datasets against different lookahead windows.
Each value, obtained by averaging the AUROC of different cross-validation folds, is plotted with its standard deviation.
In both cases, the Random Forest performance decreases for longer lookahead windows and better AUROC values are observed for the HDD dataset.
Figure \ref{fig:rf_auc} suggests that the Random Forest can efficiently predict SSD and HDD failures for $N \leq 2$ and $N \leq 8$ days lookahead, respectively.

\begin{figure}
    \centering
    \subfloat[SSD]{\includegraphics[width=0.5\columnwidth]{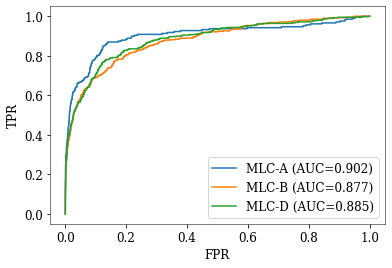}}
    \subfloat[HDD]{\includegraphics[width=0.5\columnwidth]{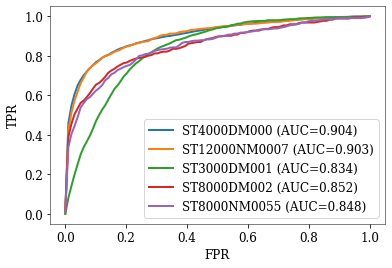}} \\
    \caption{Performance of Random Forest models when they are trained on all drive models and tested on a single one. The lookahead window is set to $N=0$ days.}
    \label{fig:single_model}
\end{figure}

Besides testing the predictor on all drive models, we also evaluate the performance of the random forest when the test set is made only of one type of drive (i.e., a single model).
Results, for lookahead window set to $N=0$ days, are shown in Figure \ref{fig:single_model} for SSD and HDD datasets.
When applied to the SSD dataset, random forests can efficiently predict the status of the SSD independently of the drive model.
Instead, in the HDD case, the predictor provides good AUROC for \textit{ST4000DM000} and \textit{ST12000NM0007}, acceptable predictions for \textit{ST8000DM002} and \textit{ST8000NM0055}, and mild performance for \textit{ST3000DM001}.

Tables \ref{tab:robust_hdd} and \ref{tab:robust_ssd} show the robustness of the Random Forest across HDD and SSD models, respectively.
For this purpose, they report the AUROC of the predictor when it is trained with a given drive model and it is tested on a different one.
While the random forest is very robust when applied to the SSD dataset, it presents some inaccuracy when trained on the HDD workload.
The worst AUROC is observed for \textit{ST3000DM001}, independently of the HDD model used for training. Good prediction performance for this HDD model is observed only when the random forest is trained using the same drive model.
This is consistent with what is observed in Tables \ref{table:different-error} and \ref{table:number-failures} that show different error and failure rates for \textit{ST3000DM001} comparing to other HDD models.
Training the random forest with all available models improves the performance of the classifier for both SSDs and HDDs.

\begin{table*}[t]
    \centering
    \caption{Random forest on the HDD dataset for $N=0$.}
    \label{tab:robust_hdd}
    \begin{tabular}{c|ccccc|c}
        \hline
        & \multicolumn{6}{c}{Training} \\
        Test & ST4000DM000 & ST12000NM0007 & ST3000DM001 & ST8000DM002 & ST8000NM0055 & All  \\
        \hline
        ST4000DM000 & 0.902 & 0.839 & 0.828 & 0.827 & 0.788 & 0.904 \\
        ST12000NM0007 & 0.872 & 0.901 & 0.828 & 0.878 & 0.848 & 0.903 \\
        ST3000DM001 & 0.745 & 0.694 & 0.863 & 0.664 & 0.654 & 0.834 \\
        ST8000DM002 & 0.809 & 0.822 & 0.774 & 0.844 & 0.832 & 0.852 \\
        ST8000NM0055 & 0.793 & 0.818 & 0.743 & 0.831 & 0.850 & 0.848 \\
        \hline 
    \end{tabular}
\end{table*}

\begin{table}[t]
    \centering
    \caption{Random forest on the SSD dataset for $N=0$.}
    \label{tab:robust_ssd}
    \begin{tabular}{c|ccc|c}
        \hline
        & \multicolumn{4}{c}{Training} \\
        Test & MLC-A & MLC-B & MLC-D & All \\
        \hline
        MLC-A & 0.891 & 0.871 & 0.887 & 0.901 \\
        MLC-B & 0.832 & 0.892 & 0.849 & 0.893 \\
        MLC-D & 0.868 & 0.857 & 0.863 & 0.901 \\
        \hline 
    \end{tabular}
\end{table}

\subsection{Model Improvement}
\label{subsec:improvements}
Section \ref{sec:causes} shows that many SSD failures are related to the drive age, while HDD ones are affected by the head flying hours (i.e., SMART 240).
Here, we use those attributes to improve the performance of the model.
Each dataset is split based on the value of the considered feature (i.e., drive age or head flying hours).
Then, the model is trained and validated on each sub-dataset and the performance of each new model is compared to the performance obtained without splitting the dataset.

\begin{figure}
    \centering
    \subfloat[SSD, Drive Age]{\label{subfig:ssd_age}\includegraphics[width=0.25\textwidth]{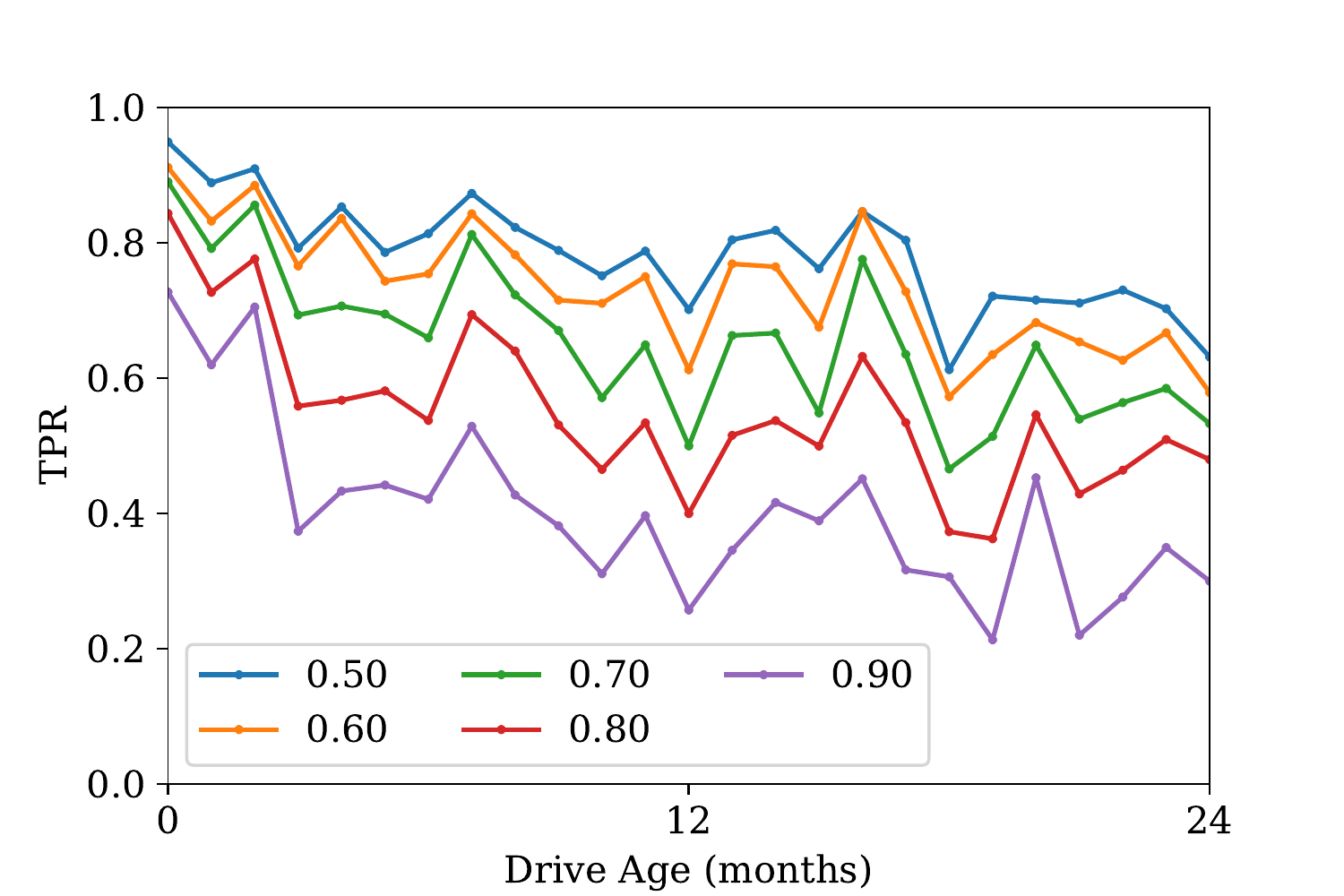}}
    \subfloat[HDD, Head Flying Hours]{\label{subfig:hdd_head}\includegraphics[width=0.25\textwidth]{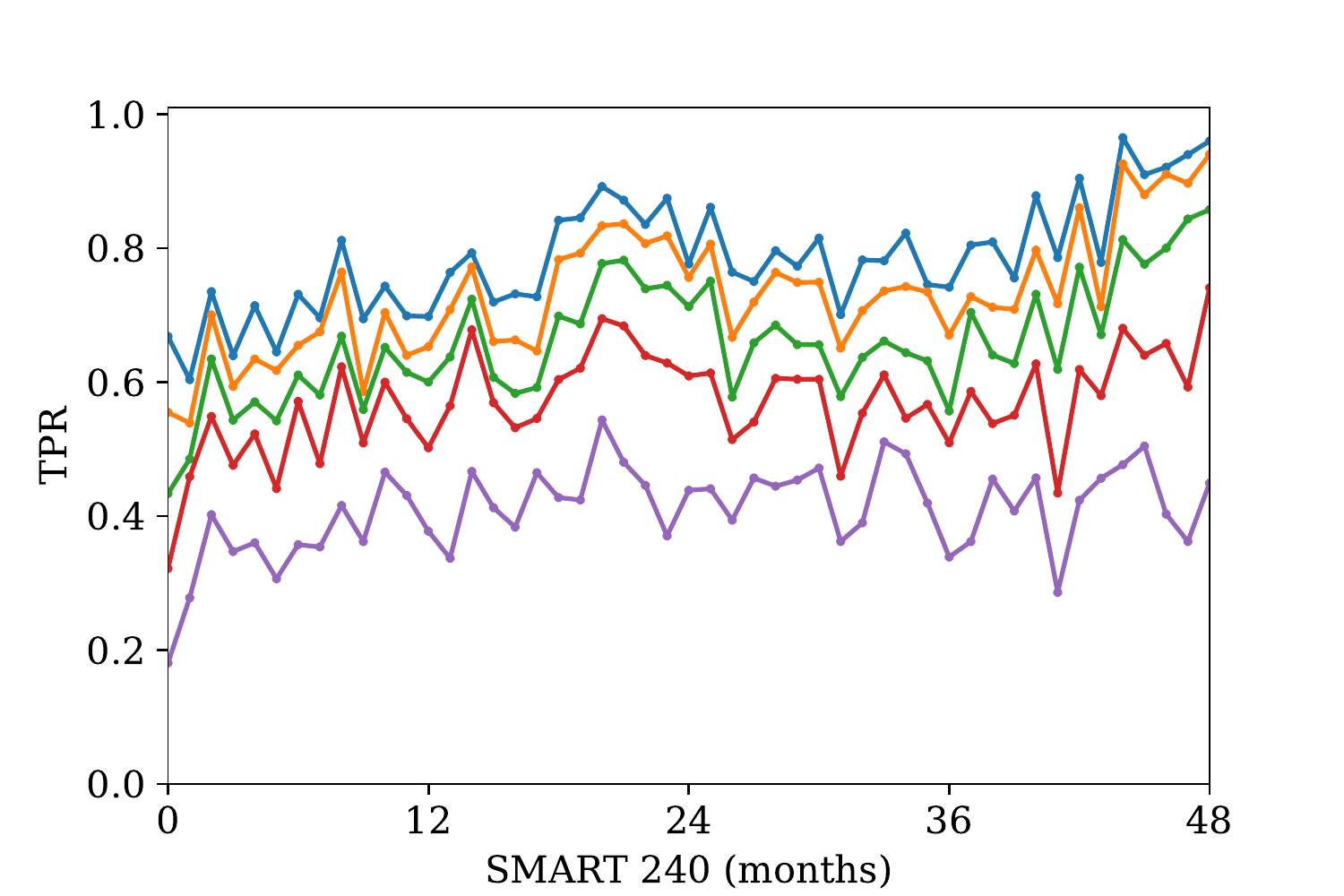}}
    \caption{True positive rate as a function of drive attributes for SSD and HDD. The lookahead window is set to N=0 days and different values of the prediction threshold, $\alpha$, are shown.}
    \label{fig:threshold}
\end{figure}

First, we evaluate the TPR of the model against different splitting on drive age (for SSDs) and head flying hours (for HDDs).
Since Random Forest models return a prediction probability, a threshold $\alpha$ is set to obtain a binary prediction.
A conservative threshold (i.e., $\alpha \simeq 1$) is practical for problems that require a low false positive rate.
We test different thresholds between 0.5 (the default one) and 1.0.
Results are shown in Fig. \ref{fig:threshold}.
Fig. \ref{fig:threshold}\subref{subfig:ssd_age} shows that, independently of the chosen threshold, the TPR for young SSDs (i.e., younger than 3 months) is substantially greater than the TPR for older drives.
Fig. \ref{fig:threshold}\subref{subfig:hdd_head} depicts the effect of the time (in months) spent for positioning the disk heads (i.e., SMART 240) on the performance of the model.
Especially for small values of $\alpha$, the TPR increases with the time spent by the drive for positioning its heads.

\begin{figure}
    \centering
    \subfloat[SSD]{\label{subfig:ssd_roc}\includegraphics[width=0.5\columnwidth]{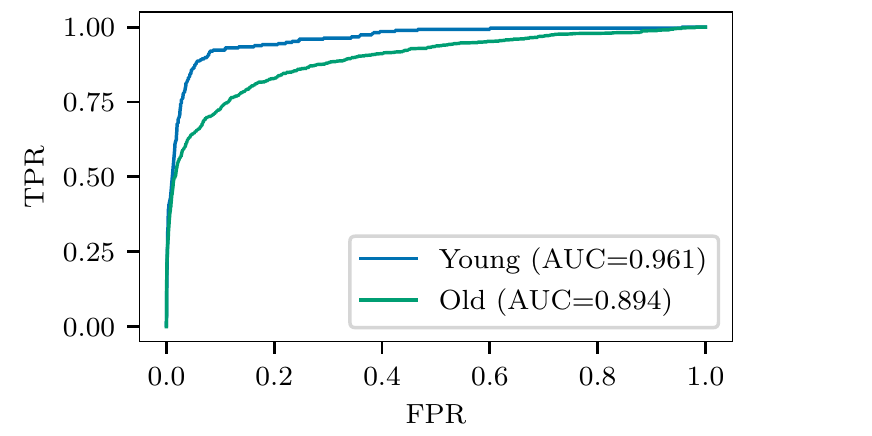}}
    \subfloat[HDD]{\label{subfig:hdd_roc}\includegraphics[width=0.5\columnwidth]{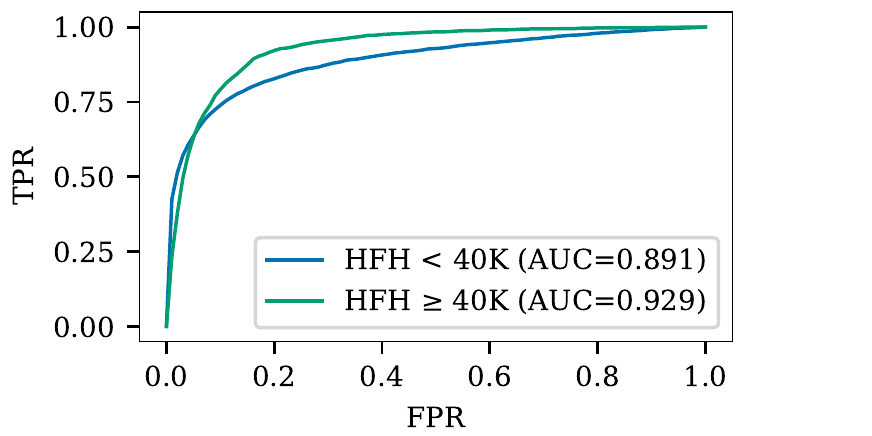}} \\
    \caption{ROC curves after splitting the dataset on drive features. Prediction model is random forest with lookahead window $N=0$ days. In \subref{subfig:hdd_roc}, \textit{HFH} is for \textit{Head Flying Hours} (i.e., SMART 240).}
    \label{fig:roc_split}
\end{figure}

Fig. \ref{fig:roc_split} shows the ROC curve of the model when it is trained on different sub-dataset and predicts the state of each drive in the next 24 hours.
As depicted in Fig. \ref{fig:roc_split}\subref{subfig:ssd_roc}, the prediction model works better with young SSDs (i.e., drive age smaller than 3 months) since its AUROC is significantly larger (0.961) than the one shown in Fig. \ref{fig:rf_auc} (0.906).
This comes at the expense of slightly reduced performance for older drives (0.894).
Similarly, Fig. \ref{fig:roc_split}\subref{subfig:hdd_roc} shows the performance improvement observed by splitting the HDD dataset on the \textit{head flying hours} feature (i.e., 40,000 hours).
The model can better predict the state of HDDs that spend a longer time in positioning their heads (0.929).
Also in this case, improvements are observed comparing to the default strategy (i.e., no split) when the measured AUROC is 0.902, while the performance for drives with small head flying hours slightly decreases (0.890).
It is worth noting that the 20\% of swap-inducing failures in the SSD dataset are young failures, while the 25\% of HDDs has a large head flying hours.


\subsection{Model Interpretability}
\label{subsec:interpretation}
Besides providing the best performance as shown in Table \ref{tab:predictors_auc}, Random Forest models also assign a score to each attribute based on its relevance for solving the classification problem.
This greatly increases the model interpretability since it is possible to identify those features that are more related to drive failures.

\begin{figure*}
    \centering
    \subfloat[SSD, split on drive age]{\label{subfig:ssd_imp}\includegraphics[width=0.4\textwidth]{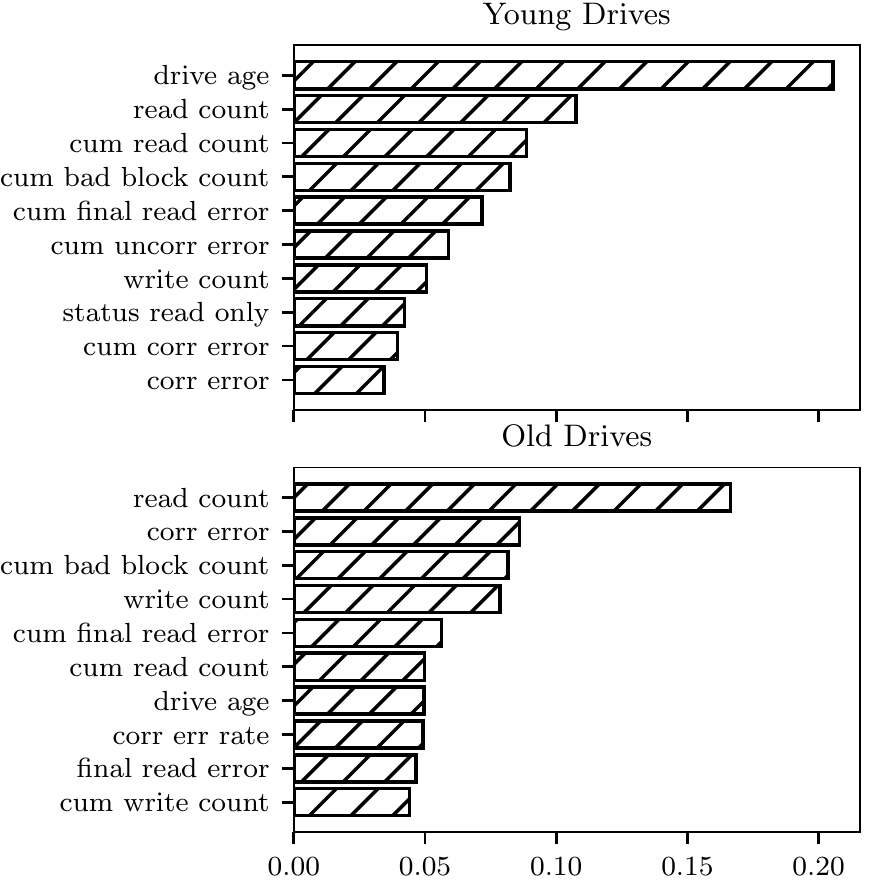}}
    \subfloat[HDD, split on head flying hours]{\label{subfig:hdd_imp}\includegraphics[width=0.4\textwidth]{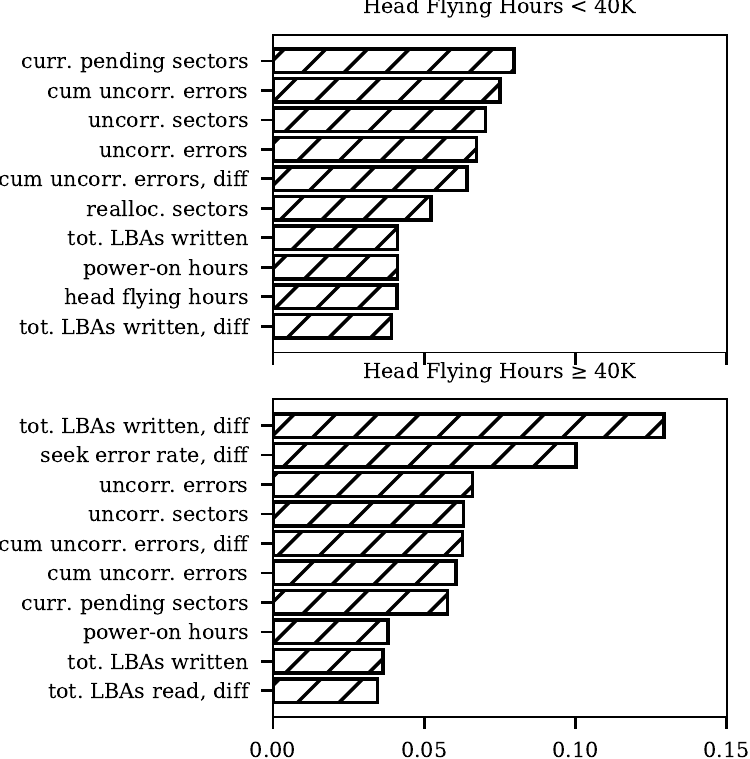}} \\
    \caption{Feature importance for Random Forest models after splitting the dataset on drive features.}
    \label{fig:imp_split}
\end{figure*}

Fig. \ref{fig:imp_split} shows the TOP-10 features for each sub-dataset considered in Section \ref{subsec:improvements} (i.e., young and old SSDs, HDDs with short and large head flying hours).
Fig. \ref{fig:imp_split}\subref{subfig:ssd_imp} shows the feature ranking for the SSD dataset.
When considering young drives, the drive age is the most important feature, followed by the read count, its cumulative value, and the cumulative number of bad blocks.
For old SSDs, features counting correctable errors and read/write operations, and the cumulative number of bad blocks are in the TOP-4.
It is expected that read and write counts are more relevant for the state prediction of old drives since young drives may have only a few activities at the failure time.
Fig. \ref{fig:imp_split}\subref{subfig:hdd_imp} depicts the feature importance for the HDD dataset.
The number of \textit{current pending sectors}, \textit{uncorrectable errors}, \textit{uncorrectable sectors}, and \textit{reallocated sectors} are among the most important features for detecting failing drives. This is similar to what is observed in \cite{mahdisoltani2017proactive}.
The attribute ranking of HDDs with large head flying hours provides new insights.
In this case, the two most relevant features are the incremental step of \textit{written logical block addressing} (LBA) and \textit{seek error rate}, followed by the number of \textit{uncorrectable errors} and \textit{uncorrectable sectors}. The \textit{seek error rate} and the \textit{uncorrectable sector} count are observed to be important features also in \cite{lu2020making}. The \textit{reallocated sector} count is not in the TOP-10 important features for HDDs with large head flying hours.

\section{Related Work}
\label{sec:related}

Large-scale data centers serve millions of requests every day \cite{wang2010impact,gill2011understanding,verma2015large} and their dependability is critical for services hosted on their infrastructure.
Much prior work investigates the main components that affect the data center dependability \cite{guo2015pingmesh,ma2015raidshield,schroeder2017reliability,wang2017can,nie2018dsn,DBLP:conf/mascots/NieXGEST17} and storage systems are pointed out as extremely important components to determine the data center reliability \cite{schroeder2007disk,schroeder2009dram,hwang2012cosmic,xu2019lessons}.

Storage drives, such as HDDs and SSDs, include a monitor system (originally designed by IBM and called SMART) that allows logging data about the drive reliability \cite{SMART}.
Disk failures are investigated by Schroeder and Gibson in \cite{schroeder2007disk}, while Pinheiro et al. \cite{pinheiro2007failure} explore their relationship with SMART features.
Meza et al. \cite{MezaWKM15} conduct a large-scale reliability study of SSDs mainly focusing on hardware events. Maneas et al. \cite{maneas2020study} explore reasons for breakdown replacement and correlations among drives in the same RAID group.
Jaffer et al. \cite{jaffer2019evaluating} evaluate the ability of different file systems to get through SSD faults.
Hao et al. \cite{hao2016tail} empirically study HDDs and SSDs in the field to highlight the importance of masking storage tail latencies to increase performance stability.
None of these works attempts to predict the failure state of drives.

Different approaches are proposed to predict drives failures.
Hamerly et al. \cite{hamerly2001bayesian} and Hughes et al. \cite{hughes2002improved} use statistical inference (i.e., Bayesian approach and Rank sum hypothesis test, respectively), while Ma et al. \cite{ma2015raidshield} rely on thresholds to predict component failures.
Goldszmidt \cite{goldszmidt2012finding} combines statistical models and thresholds to predict disks that are soon to fail.
However, machine learning prediction models are shown to be more efficient than other strategies.
Murray et al. \cite{murray2005machine} develop a multi-instance naive Bayes classifier to reduce the number of false alarms when predicting disk failures, while Tan and Gu \cite{tan2010predictability} investigates the performance of a tree augmented naive Bayesian method to predict the future drive status.
Agarwal et al. \cite{agarwal2009discovering} investigate the performance of a rule-based classifier for discovering disk failures, Li et al. \cite{li2014hard} address the same problem by using decision trees, and Zhu et al. \cite{zhu2013proactive} adopt neural networks and SVM.
Lu et al. \cite{lu2020making} implement a convolutional neural network with long short-term memory that predicts HDD failures with a 10-day lookahead window by considering SMART features, disk performance metrics, and disk physical location.
Queiroz et al. \cite{queiroz2017fault} propose a new method (i.e., Gaussian Mixture based Fault Detection) to build a statistical model of SMART attributes and predict imminent HDD failures.

To fairly compare the performance of our method with other approaches, the same dataset must be used.
As discussed by Yu \cite{yu2019hard}, the performance may be extremely different when considering different data sources.
The Backblaze dataset is used in \cite{botezatu2016predicting,aussel2017predictive} to train and validate their approaches for predicting disk failures while \cite{mahdisoltani2017proactive} uses the Backblaze and the Google datasets for investigating faulty HDDs and SSDs.
The regularized greedy forest adopted by Botezatu et al. \cite{botezatu2016predicting} shines when compared to other approaches used to predict failures in the Backblaze dataset with 98\% precision and accuracy. Differently from our approach, \cite{botezatu2016predicting} splits observations from the same drive into training and test sets not considering error and workload correlations across different drive days.
It also undersamples both training and test sets \cite{aussel2017predictive}.
Similarly, also Wang et al. \cite{wang2019attention} test a deep architecture for predicting HDD failures on an undersampled dataset.
However, resampling the whole dataset (training and test sets) generates overoptimistic results.
Aussel et al. \cite{aussel2017predictive} train and evaluate a random forest on a small subset of the Backblaze dataset (i.e., only data from 2014).
Although they show high precision and a considerable recall (i.e., 95\% and 65\%, respectively), they filter out observations with similar features and different failure states.
We believe this is hardly doable in real scenarios since it requires an apriori knowledge of the drive state.
Mahdisoltani et al. \cite{mahdisoltani2017proactive} investigates different prediction models to predict uncorrectable errors and bad blocks in HDDs and SSDs, and they show that random forests provide good performance for this task.
Although our approach is similar to the one presented in \cite{mahdisoltani2017proactive}, we aim to predict drive failures and do not try to predict errors.
Approaches for online HDD failure prediction are also investigated \cite{xiao2018disk,han2020toward} by using online random forests and extending available machine learning models with concept drift adaptation \cite{gama2014survey}.
Unfortunately, the SSD dataset does not provide a global timestamp, hence online prediction cannot be implemented here.

In this paper, we explore the capability of random forests for predicting drive failures and investigate possible enhancements by statistically analyzing drive features and using different models based on observed attributes.
We consider a conceivably long lookahead window and use two large and real datasets for training and validating the proposed machine learning approach.
To the best of our knowledge, storage device failures have never been studied by splitting the dataset based on attribute values.

\section{Conclusion}
\label{sec:conclusion}

In this paper, we investigate SSD and HDD failures using two traces, each one six-year long, from production environments.
Daily logs for 30,000 SSDs are collected at a Google data center, while 100,000 HDDs are observed at a Backblaze data center.
Analyzing the available traces, we draw remarkable conclusions.
Unexpectedly, we observe that features that are commonly thought to cause SSD failures (i.e., write operations and error incidence) are not highly related to faulty SSDs.
We train and test different classifiers to predict faulty SSDs and HDDs, and note that Random Forest models provide accurate predictions with a short training time.
Their high interpretability makes them the best predictor for the analyzed problem.
We observe that splitting each dataset based on attribute values of its observations allows increasing the performance of random forests.
The drive age is a critical attribute for predicting SSD failures; drives failing before being three months old can be detected easier than other drives.
Similarly, when predicting faulty HDDs, a higer detection rate is observed for drives with \textit{head flying hours} (i.e., SMART 240) longer than 40,000 hours.

{\small
\printbibliography}

@article{fawcett2006introduction,
  title={An introduction to ROC analysis},
  author={Fawcett, Tom},
  journal={Pattern recognition letters},
  volume={27},
  number={8},
  pages={861--874},
  year={2006},
  publisher={Elsevier}
}

@inproceedings{alter2019ssd,
  title={SSD failures in the field: symptoms, causes, and prediction models},
  author={Alter, Jacob and Xue, Ji and Dimnaku, Alma and Smirni, Evgenia},
  booktitle={Proceedings of the International Conference for High Performance Computing, Networking, Storage and Analysis},
  pages={1--14},
  year={2019}
}

@inproceedings{botezatu2016predicting,
  title={Predicting disk replacement towards reliable data centers},
  author={Botezatu, Mirela Madalina and Giurgiu, Ioana and Bogojeska, Jasmina and Wiesmann, Dorothea},
  booktitle={Proceedings of the 22nd ACM SIGKDD International Conference on Knowledge Discovery and Data Mining},
  pages={39--48},
  year={2016}
}

@inproceedings{goldszmidt2012finding,
  author    = {Mois{\'{e}}s Goldszmidt},
  title     = {Finding Soon-to-Fail Disks in a Haystack},
  booktitle = {4th {USENIX} Workshop on Hot Topics in Storage and File Systems, HotStorage'12},
  year      = {2012}
}

@inproceedings{hamerly2001bayesian,
  title={Bayesian approaches to failure prediction for disk drives},
  author={Hamerly, Greg and Elkan, Charles and others},
  booktitle={ICML},
  volume={1},
  pages={202--209},
  year={2001}
}

@inproceedings{pinheiro2007failure,
  title={Failure trends in a large disk drive population},
  author={Pinheiro, Eduardo and Weber, Wolf-Dietrich and Barroso, Luiz Andr{\'e}},
  booktitle={5th {USENIX} Conference on File and Storage Technologies, {FAST} 2007},
  pages={17--28},
  year={2007}
}

@article{hughes2002improved,
  title={Improved disk-drive failure warnings},
  author={Hughes, Gordon F and Murray, Joseph F and Kreutz-Delgado, Kenneth and Elkan, Charles},
  journal={IEEE transactions on reliability},
  volume={51},
  number={3},
  pages={350--357},
  year={2002},
  publisher={IEEE}
}

@inproceedings{li2014hard,
  title={Hard drive failure prediction using classification and regression trees},
  author={Li, Jing and Ji, Xinpu and Jia, Yuhan and Zhu, Bingpeng and Wang, Gang and Li, Zhongwei and Liu, Xiaoguang},
  booktitle={2014 44th Annual IEEE/IFIP International Conference on Dependable Systems and Networks},
  pages={383--394},
  year={2014},
  organization={IEEE}
}

@article{ma2015raidshield,
  title={RAIDShield: characterizing, monitoring, and proactively protecting against disk failures},
  author={Ma, Ao and Traylor, Rachel and Douglis, Fred and Chamness, Mark and Lu, Guanlin and Sawyer, Darren and Chandra, Surendar and Hsu, Windsor},
  journal={ACM Transactions on Storage (TOS)},
  volume={11},
  number={4},
  pages={1--28},
  year={2015},
  publisher={ACM New York, NY, USA}
}

@article{murray2005machine,
  title={Machine learning methods for predicting failures in hard drives: A multiple-instance application},
  author={Murray, Joseph F and Hughes, Gordon F and Kreutz-Delgado, Kenneth},
  journal={Journal of Machine Learning Research},
  volume={6},
  pages={783--816},
  year={2005}
}

@inproceedings{zhu2013proactive,
  title={Proactive drive failure prediction for large scale storage systems},
  author={Zhu, Bingpeng and Wang, Gang and Liu, Xiaoguang and Hu, Dianming and Lin, Sheng and Ma, Jingwei},
  booktitle={2013 IEEE 29th symposium on mass storage systems and technologies (MSST)},
  pages={1--5},
  year={2013},
  organization={IEEE}
}

@inproceedings{agarwal2009discovering,
  title={Discovering rules from disk events for predicting hard drive failures},
  author={Agarwal, Vipul and Bhattacharyya, Chiranjib and Niranjan, Thirumale and Susarla, Sai},
  booktitle={2009 International Conference on Machine Learning and Applications (ICMLA)},
  pages={782--786},
  year={2009},
  organization={IEEE}
}

@inproceedings{schroeder2007disk,
  title={Disk failures in the real world: What does an MTTF of 1,000,000 hours mean to you?},
  author={Schroeder, Bianca and Gibson, Garth A},
  booktitle={5th USENIX Conference on File and Storage Technologies, {FAST} 2007},
  pages={1--16},
  year={2007}
}

@inproceedings{tan2010predictability,
  title={On predictability of system anomalies in real world},
  author={Tan, Yongmin and Gu, Xiaohui},
  booktitle={2010 IEEE International Symposium on Modeling, Analysis and Simulation of Computer and Telecommunication Systems},
  pages={133--140},
  year={2010},
  organization={IEEE}
}

@inproceedings{yu2019hard,
  title={Hard disk Drive Failure Prediction Challenges in Machine Learning for Multi-variate Time Series},
  author={Yu, Jie},
  booktitle={Proceedings of the 2019 3rd International Conference on Advances in Image Processing},
  pages={144--148},
  year={2019}
}

@inproceedings{aussel2017predictive,
  title={Predictive models of hard drive failures based on operational data},
  author={Aussel, Nicolas and Jaulin, Samuel and Gandon, Guillaume and Petetin, Yohan and Fazli, Eriza and Chabridon, Sophie},
  booktitle={2017 16th IEEE International Conference on Machine Learning and Applications (ICMLA)},
  pages={619--625},
  year={2017},
  organization={IEEE}
}

@inproceedings{verma2015large,
  title={Large-scale cluster management at {G}oogle with {B}org},
  author={Verma, Abhishek and Pedrosa, Luis and Korupolu, Madhukar and Oppenheimer, David and Tune, Eric and Wilkes, John},
  booktitle={Proceedings of the Tenth European Conference on Computer Systems, EuroSys 2015},
  pages={18:1--18:17},
  year={2015},
  organization={ACM}
}

@inproceedings{gill2011understanding,
  title={Understanding network failures in data centers: measurement, analysis, and implications},
  author={Gill, Phillipa and Jain, Navendu and Nagappan, Nachiappan},
  booktitle={ACM SIGCOMM Computer Communication Review},
  volume={41},
  number={4},
  pages={350--361},
  year={2011},
  organization={ACM}
}

@inproceedings{wang2010impact,
  title={The impact of virtualization on network performance of {A}mazon {EC2} data center},
  author={Wang, Guohui and Ng, TS Eugene},
  booktitle={Infocom, 2010 proceedings ieee},
  pages={1--9},
  year={2010},
  organization={IEEE}
}

@inproceedings{wang2017can,
  title={What Can We Learn from Four Years of Data Center Hardware Failures?},
  author={Wang, Guosai and Zhang, Lifei and Xu, Wei},
  booktitle={Dependable Systems and Networks (DSN), 2017 47th Annual IEEE/IFIP International Conference on},
  pages={25--36},
  year={2017},
  organization={IEEE}
}

@article{schroeder2017reliability,
  title={Reliability of {NAND}-based {SSDs}: What field studies tell us},
  author={Schroeder, Bianca and Merchant, Arif and Lagisetty, Raghav},
  journal={Proceedings of the IEEE},
  volume={105},
  number={9},
  pages={1751--1769},
  year={2017},
  publisher={IEEE}
}

@inproceedings{schroeder2009dram,
  title={{DRAM} errors in the wild: a large-scale field study},
  author={Schroeder, Bianca and Pinheiro, Eduardo and Weber, Wolf-Dietrich},
  booktitle={ACM SIGMETRICS Performance Evaluation Review},
  volume={37},
  number={1},
  pages={193--204},
  year={2009},
  organization={ACM}
}

@inproceedings{nie2018dsn,
  title={Machine Learning Models for {GPU} Error Prediction in a Large Scale {HPC} System},
  author={Nie, Bin and Xue, Ji and Gupta, Saurabh and Patel, Tirthak and Engelmann, Christian and Smirni, Evgenia and Tiwari, Devesh},
  booktitle={Dependable Systems and Networks (DSN), 2018 48th Annual IEEE/IFIP International Conference on},
  pages={1--12},
  year={2018},
  organization={IEEE}
}

@inproceedings{mahdisoltani2017proactive,
  title={Proactive error prediction to improve storage system reliability},
  author={Mahdisoltani, Farzaneh and Stefanovici, Ioan and Schroeder, Bianca},
  booktitle={2017 USENIX Annual Technical Conference (USENIX ATC 17)},
  pages={391--402},
  year={2017}
}

@inproceedings{hwang2012cosmic,
  title={Cosmic rays don't strike twice: understanding the nature of {DRAM} errors and the implications for system design},
  author={Hwang, Andy A and Stefanovici, Ioan A and Schroeder, Bianca},
  booktitle={ACM SIGPLAN Notices},
  volume={47},
  number={4},
  pages={111--122},
  year={2012},
  organization={ACM}
}

@inproceedings{guo2015pingmesh,
  title={Pingmesh: A large-scale system for data center network latency measurement and analysis},
  author={Guo, Chuanxiong and Yuan, Lihua and Xiang, Dong and Dang, Yingnong and Huang, Ray and Maltz, Dave and Liu, Zhaoyi and Wang, Vin and Pang, Bin and Chen, Hua and others},
  booktitle={ACM SIGCOMM Computer Communication Review},
  volume={45},
  number={4},
  pages={139--152},
  year={2015},
  organization={ACM}
}

@inproceedings{MezaWKM15,
  author    = {Justin Meza and
               Qiang Wu and
               Sanjeev Kumar and
               Onur Mutlu},
  title     = {A Large-Scale Study of Flash Memory Failures in the Field},
  booktitle = {Proceedings of the 2015 {ACM} {SIGMETRICS} International Conference
               on Measurement and Modeling of Computer Systems},
  pages     = {177--190},
  year      = {2015},
  bibsource = {dblp computer science bibliography, https://dblp.org}
}

@inproceedings{BiancaFAST16,
  author    = {Bianca Schroeder and
               Raghav Lagisetty and
               Arif Merchant},
  title     = {Flash Reliability in Production: the Expected and the Unexpected},
  booktitle = {14th {USENIX} Conference on File and Storage Technologies, {FAST}
               2016},
  pages     = {67--80},
  year      = {2016},
}

@inproceedings{XuATC18,
  author    = {Yong Xu and
               Kaixin Sui and
               Randolph Yao and
               Hongyu Zhang and
               Qingwei Lin and
               Yingnong Dang and
               Peng Li and
               Keceng Jiang and
               Wenchi Zhang and
               Jian{-}Guang Lou and
               Murali Chintalapati and
               Dongmei Zhang},
  title     = {Improving Service Availability of Cloud Systems by Predicting Disk
               Error},
  booktitle = {2018 {USENIX} Annual Technical Conference, ({USENIX} {ATC} 18)},
  pages     = {481--494},
  year      = {2018}
}

@inproceedings{DBLP:conf/sigmetrics/LuoGCHM18,
  author    = {Yixin Luo and
               Saugata Ghose and
               Yu Cai and
               Erich F. Haratsch and
               Onur Mutlu},
  title     = {Improving {3D} {NAND} Flash Memory Lifetime by Tolerating Early Retention
               Loss and Process Variation},
  booktitle = {Abstracts of the 2018 {ACM} International Conference on Measurement
               and Modeling of Computer Systems, {SIGMETRICS} 2018},
  pages     = {106},
  year      = {2018}
}

@inproceedings{DBLP:conf/hpca/CaiLHMM15,
  author    = {Yu Cai and
               Yixin Luo and
               Erich F. Haratsch and
               Ken Mai and
               Onur Mutlu},
  title     = {Data retention in {MLC} {NAND} flash memory: Characterization, optimization,
               and recovery},
  booktitle = {21st {IEEE} International Symposium on High Performance Computer Architecture,
               {HPCA} 2015},
  pages     = {551--563},
  year      = {2015}
}

@inproceedings{DBLP:conf/micro/GruppCCSYSW09,
  author    = {Laura M. Grupp and
               Adrian M. Caulfield and
               Joel Coburn and
               Steven Swanson and
               Eitan Yaakobi and
               Paul H. Siegel and
               Jack K. Wolf},
  title     = {Characterizing flash memory: anomalies, observations, and applications},
  booktitle = {42st Annual {IEEE/ACM} International Symposium on Microarchitecture {(MICRO-42} 2009)},
  pages     = {24--33},
  year      = {2009}
}

@inproceedings{DBLP:conf/fast/ZhengTQL13,
  author    = {Mai Zheng and
               Joseph Tucek and
               Feng Qin and
               Mark Lillibridge},
  title     = {Understanding the robustness of {SSDs} under power fault},
  booktitle = {Proceedings of the 11th {USENIX} conference on File and Storage Technologies,
               {FAST} 2013},
  pages     = {271--284},
  year      = {2013}
}

@inproceedings{DBLP:conf/dsn/QureshiKKNM15,
  author    = {Moinuddin K. Qureshi and
               Dae{-}Hyun Kim and
               Samira Manabi Khan and
               Prashant J. Nair and
               Onur Mutlu},
  title     = {{AVATAR:} A Variable-Retention-Time {(VRT)} Aware Refresh for {DRAM}
               Systems},
  booktitle = {45th Annual {IEEE/IFIP} International Conference on Dependable Systems
               and Networks, {DSN} 2015},
  pages     = {427--437},
  year      = {2015},
}

@misc{SMART,
  title = {American {N}ational {S}tandards {I}nstitute. {AT attachment 8 - ATA/ATAPI} command set {(ATA8-ACS)}, 2008.},
  howpublished = {\url{http://www.t13.org/documents/uploadeddocuments/docs2008/d1699r6a-ata8-acs.pdf}},
  note = {[Online; accessed on 2020-04-19]}
}

@BOOK{statistics,
  TITLE = {Nonparametric {S}tatistics: {A Step-by-Step Approach}},
  AUTHOR = {Corder, G.W. and Foreman, D.I.},
  YEAR = {2014},
  PUBLISHER = {Wiley}
}

@article{wang2019attention,
  title={An Attention-augmented Deep Architecture for Hard Drive Status Monitoring in Large-scale Storage Systems},
  author={Wang, Ji and Bao, Weidong and Zheng, Lei and Zhu, Xiaomin and Yu, Philip S},
  journal={ACM Transactions on Storage (TOS)},
  volume={15},
  number={3},
  pages={1--26},
  year={2019},
  publisher={ACM New York, NY, USA}
}

@inproceedings{jaffer2019evaluating,
  title={Evaluating file system reliability on solid state drives},
  author={Jaffer, Shehbaz and Maneas, Stathis and Hwang, Andy and Schroeder, Bianca},
  booktitle={2019 USENIX Annual Technical Conference (USENIX ATC 19)},
  pages={783--798},
  year={2019}
}

@inproceedings{xu2019lessons,
  title={Lessons and actions: What we learned from 10k ssd-related storage system failures},
  author={Xu, Erci and Zheng, Mai and Qin, Feng and Xu, Yikang and Wu, Jiesheng},
  booktitle={2019 USENIX Annual Technical Conference (USENIX ATC 19)},
  pages={961--976},
  year={2019}
}

@misc{backblaze,
  title={Hard Drive Data and Stats},
  author={Backblaze},
  howpublished={\url{https://www.backblaze.com/b2/hard-drive-test-data.html}},
  note={[Online; accessed on 2020-04-28]}
}

@inproceedings{xiao2018disk,
  title={Disk failure prediction in data centers via online learning},
  author={Xiao, Jiang and Xiong, Zhuang and Wu, Song and Yi, Yusheng and Jin, Hai and Hu, Kan},
  booktitle={Proceedings of the 47th International Conference on Parallel Processing},
  pages={1--10},
  year={2018}
}

@inproceedings{han2020toward,
  title={Toward Adaptive Disk Failure Prediction via Stream Mining},
  author={Han, Shujie and Lee, Patrick PC and Shen, Zhirong and He, Cheng and Liu, Yi and Huang, Tao},
  booktitle={2019 IEEE 40th International Conference on Distributed Computing Systems (ICDCS)},
  %pages={xx--yy},
  year={2020},
  organization={IEEE}
}

@article{gama2014survey,
  title={A survey on concept drift adaptation},
  author={Gama, Jo{\~a}o and {\v{Z}}liobait{\.e}, Indr{\.e} and Bifet, Albert and Pechenizkiy, Mykola and Bouchachia, Abdelhamid},
  journal={ACM computing surveys (CSUR)},
  volume={46},
  number={4},
  pages={1--37},
  year={2014}
}

@inproceedings{lu2020making,
  title={Making Disk Failure Predictions SMARTer!},
  author={Lu, Sidi and Luo, Bing and Patel, Tirthak and Yao, Yongtao and Tiwari, Devesh and Shi, Weisong},
  booktitle={18th USENIX Conference on File and Storage Technologies (FAST 20)},
  pages={151--167},
  year={2020}
}

@inproceedings{hao2016tail,
  title={The Tail at Store: A Revelation from Millions of Hours of Disk and SSD Deployments},
  author={Hao, Mingzhe and Soundararajan, Gokul and Kenchammana-Hosekote, Deepak and Chien, Andrew A and Gunawi, Haryadi S},
  booktitle={14th USENIX Conference on File and Storage Technologies (FAST 16)},
  pages={263--276},
  year={2016}
}

@article{queiroz2017fault,
  title={A fault detection method for hard disk drives based on mixture of Gaussians and nonparametric statistics},
  author={Queiroz, Lucas P and Rodrigues, Francisco Caio M and Gomes, Joao Paulo P and Brito, Felipe T and Chaves, Iago C and Paula, Manoel Rui P and Salvador, Marcos R and Machado, Javam C},
  journal={IEEE Transactions on Industrial Informatics},
  volume={13},
  number={2},
  pages={542--550},
  year={2017},
  publisher={IEEE}
}

@inproceedings{DBLP:conf/fast/BirkeBCSE14,
  author    = {Robert Birke and
               Mathias Bj{\"{o}}rkqvist and
               Lydia Y. Chen and
               Evgenia Smirni and
               Ton Engbersen},
  title     = {(Big)data in a virtualized world: volume, velocity, and variety in
               cloud datacenters},
  booktitle = {Proceedings of the 12th {USENIX} conference on File and Storage Technologies,
               {FAST} 2014, Santa Clara, CA, USA, February 17-20, 2014},
  pages     = {177--189},
  publisher = {{USENIX}},
  year      = {2014}
}

@inproceedings{DBLP:conf/mascots/NieXGEST17,
  author    = {Bin Nie and
               Ji Xue and
               Saurabh Gupta and
               Christian Engelmann and
               Evgenia Smirni and
               Devesh Tiwari},
  title     = {Characterizing Temperature, Power, and Soft-Error Behaviors in Data
               Center Systems: Insights, Challenges, and Opportunities},
  booktitle = {25th {IEEE} International Symposium on Modeling, Analysis, and Simulation
               of Computer and Telecommunication Systems, {MASCOTS} 2017, Banff,
               AB, Canada, September 20-22, 2017},
  pages     = {22--31},
  publisher = {{IEEE} Computer Society},
  year      = {2017}
}

@inproceedings{maneas2020study,
  title={A Study of SSD Reliability in Large Scale Enterprise Storage Deployments},
  author={Maneas, Stathis and Mahdaviani, Kaveh and Emami, Tim and Schroeder, Bianca},
  booktitle={18th USENIX Conference on File and Storage Technologies (FAST 20)},
  pages={137--149},
  year={2020}
}

\end{document}